\documentclass[runningheads]{llncs}

\usepackage{eccv}

\usepackage{eccvabbrv}

\usepackage{graphicx}
\usepackage{booktabs}
\usepackage{multirow}
\usepackage{pifont}
\usepackage[normalem]{ulem}
\useunder{\uline}{\ul}{}
\usepackage[accsupp]{axessibility}  
\usepackage{color, colortbl}
\definecolor{LightCyan}{rgb}{0.88,1,1}

\usepackage{listings}
\usepackage{xcolor}

\usepackage{hyperref}

\usepackage{orcidlink}

\begin{document}

\title{Image-to-Lidar Relational Distillation for Autonomous Driving Data} 


\author{Anas Mahmoud\inst{1}\orcidlink{0000-0003-0836-7416} \and
Ali Harakeh\inst{2}\orcidlink{0000-0002-2593-2141} \and
Steven Waslander\inst{1}\orcidlink{0000-0003-4217-4415}}

\authorrunning{A.~Mahmoud et al.}

\institute{University of Toronto \\
 \and
Mila - Quebec AI Institute}

\maketitle


\begin{abstract}
    
Pre-trained on extensive and diverse multi-modal datasets, 2D foundation models excel at addressing 2D tasks with little or no downstream supervision, owing to their robust representations.
The emergence of 2D-to-3D distillation frameworks has extended these capabilities to 3D models.
    However, distilling 3D representations for autonomous driving datasets presents challenges like self-similarity, class imbalance, and point cloud sparsity, hindering the effectiveness of contrastive distillation, especially in zero-shot learning contexts. Whereas other methodologies, such as similarity-based distillation, enhance zero-shot performance, they tend to yield less discriminative representations, diminishing few-shot performance.
   We investigate the gap in structure between the 2D and the 3D representations that result from state-of-the-art distillation frameworks and reveal a significant mismatch between the two. Additionally, we demonstrate that the observed structural gap is negatively correlated with the efficacy of the distilled representations on zero-shot and few-shot 3D semantic segmentation.
   To bridge this gap, we propose a relational distillation framework enforcing intra-modal and cross-modal constraints, resulting in distilled 3D representations that closely capture the structure of the 2D representation. 
This alignment significantly enhances 3D representation performance over those learned through contrastive distillation in zero-shot segmentation tasks. Furthermore, our relational loss consistently improves the quality of 3D representations in both in-distribution and out-of-distribution few-shot segmentation tasks, outperforming approaches that rely on the similarity loss.

\end{abstract}

\section{Introduction}
\label{sec:intro}

\begin{figure}[t]
  \centering
  \includegraphics[width=\textwidth]{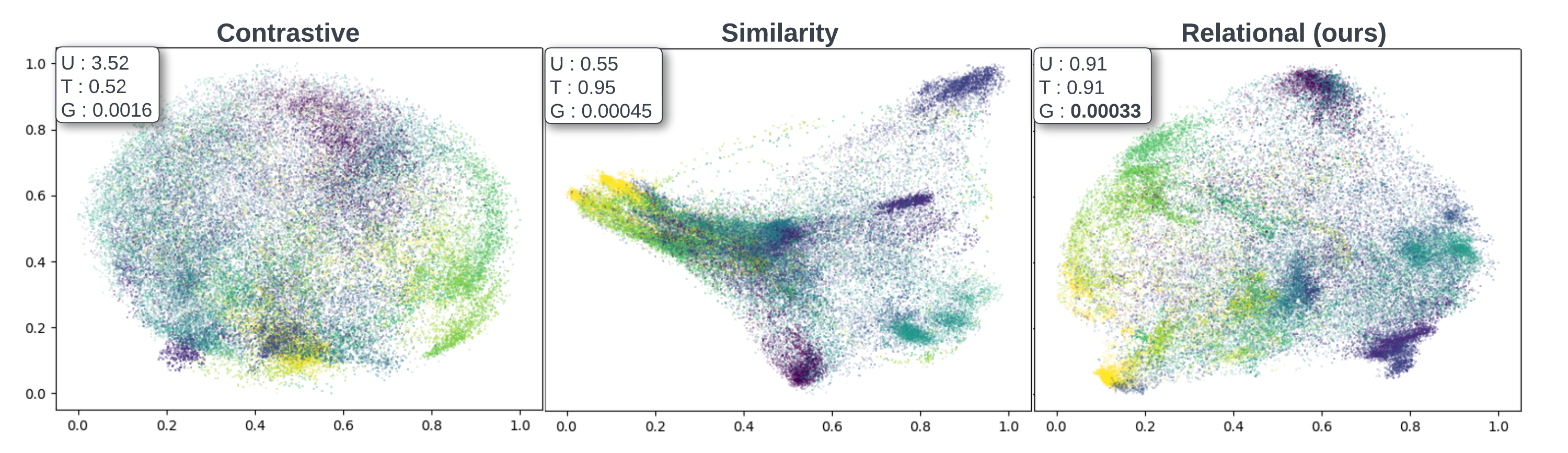}
  \caption{We distill 2D representations from CLIP~\cite{radford2021learningclip} to a 3D point-cloud encoder using the contrastive loss, similarity loss, and our proposed relational loss, and compute the uniformity (U), tolerance (T), and modality gap (G) of the learned 3D representations. We sample $5000$ point features from each of the $16$ classes defined in the nuScenes dataset~\cite{caesar2020nuscenes}, apply PCA and visualize the primary components. The source U and T of the CLIP image encoder are $1.54$ and $0.73$, respectively. Compared to the source, we see that contrastive loss learns 3D representations with higher U and lower T compared to the source, while the trends are reversed for similarity loss.  Our proposed relational loss minimizes this structural mismatch and leads to the lowest modality gap.
  }
  \label{fig:contrastive_sim_rel_pointlosses}
  \vspace{-15pt}
\end{figure}

Understanding 3D scenes is pivotal for robotics applications~\cite{conceptfusion, qian2022pocd}, notably in autonomous driving, where accurate navigation and decision-making depend on precise environmental perception. 
Solving the perception tasks necessary for 3D scene understanding requires the point-wise labelling of LiDAR scenes, which is tedious, compounded by the sparsity of LiDAR data, and costly~\cite{dvf, bounding_box_painting}. These issues result in a scarcity of large-scale, diverse point cloud datasets, particularly those aligned with images or text, significantly hampering the development of foundation models for 3D tasks. This shortage is particularly problematic for few-shot or zero-shot learning approaches, which aim to achieve proficiency with minimal or no labelled examples~\cite{parnami2022fewlearning, gpt3}. Bridging this gap is vital for creating models adept at understanding complex 3D scenes with limited data.

Contrary to the 3D domain, the 2D and language domains benefit from the availability of large-scale, diverse, and multi-modal datasets, which facilitated the development of Vision Foundation Models (VFMs)~\cite{caron2021emerging_dino, oquab2023dinov2}, and Vision-Language Models (VLMs)~\cite{radford2021learningclip, zhou2022extractMaskCLIP}. These models have shown remarkable label efficiency in 2D tasks like image classification and segmentation and have also been used to perform label-free 2D image classification and segmentation through language prompts during inference~\cite{radford2021learningclip, mahmoud2024sieve}. Due to these advantages, recent approaches for learning 3D representations have relied on the distillation of 2D representations from VFMs~\cite{liu2021ppkt, sautier2022slidr, stcl} or VLMs~\cite{zhang2022pointclip, chen2023clip2scene, peng2023openscene} to point cloud encoders, and have shown encouraging results when solving few-shot and zero-shot 3D tasks.

Unfortunately, the adaptation of 2D-to-3D distillation frameworks for autonomous driving (AD) datasets reveals unique challenges, notably due to the inherent characteristics of AD data. Current frameworks, largely reliant on contrastive learning methods, face the issue of self-similarity~\cite{stcl}, a prevalent phenomenon in AD datasets. Self-similarity arises when a significant portion of the training examples belong to a single semantic category (e.g. road, trees or sky in AD data). Under the effect of self-similarity, the contrastive loss mechanism, designed to be hardness-aware~\cite{wang2021understanding}, inadvertently pushes away semantically similar samples, which not only disrupts the local semantic coherence of the 3D representation~\cite{sautier2022slidr, stcl, liu2023seal} but also amplifies the effects of AD datasets' severe class imbalance. As a result, while such frameworks may result in useful 3D representations for few-shot learning tasks, the misalignment induced by excessive pushing of semantically similar examples undermines their efficacy in zero-shot learning scenarios, where precise cross-modal alignment is crucial.
Another approach to distillation relies on the cosine similarity loss, which attempts to learn a 3D representation by driving the features generated for every 3D point by a point encoder to its corresponding 2D feature from VFMs or VLMs.  Using the similarity loss results in 3D representations that achieve significantly better performance on the zero-shot tasks compared to ones learned with the contrastive loss. However, we argue in Section~\ref{sec:methodology} that using the similarity loss under-constrains the pretraining when compared to the contrastive loss, resulting in sub-optimal 3D representations on few-shot downstream tasks.

To highlight the mismatch discussed above, we provide an example visualization of the 3D representation space (\cref{fig:contrastive_sim_rel_pointlosses}), which was obtained by distilling from CLIP's~\cite{radford2021learningclip} 2D representation space using both the contrastive and similarity losses. We observe that both losses result in 3D representations that diverge from the structure of the 2D representation as measured by uniformity, tolerance, and the modality gap, which we further explain in Section~\ref{sec:methodology}.


In this work, we investigate the impact of state-of-the-art 2D-to-3D distillation frameworks on the structure of learned 3D representations. We show that the choice of loss during pretraining can result in a significant mismatch between the structure of the 2D source representations and the distilled 3D representations. Furthermore, we demonstrate that this mismatch leads to a deterioration in performance on downstream tasks. Our contributions are as follows:

\begin{itemize}
    \item \textbf{Quantify the Gap:} We quantify the mismatch in structure when performing distillation using contrastive loss~\cite{liu2021ppkt, sautier2022slidr, girdhar2023imagebind} and similarity loss~\cite{Hess_2024_WACV_LidarCLIP, peng2023openscene} via the uniformity~\cite{wang2021understanding, wang2020understanding_uniformity_alignment}, tolerance~\cite{wang2021understanding}, and modality gap~\cite{liang2022mind}, revealing a significant gap between 2D and distilled 3D representations.

\item \textbf{Bridging the Gap using Relational Distillation:}
We address this mismatch by imposing structural constraints that foster the learning of a 3D representation aligned with the structure of 2D representations. To achieve this, we employ pretraining with intra-modal and cross-modal relational losses. These losses generalize the similarity loss, providing a more effective constraint on the distillation process. Our proposed losses can be applied to pixel-based~\cite{liu2021ppkt} and superpixel-based~\cite{sautier2022slidr} distillation frameworks. 

\item \textbf{Bridging the Gap Improves Downstream Performance:} Our proposed loss effectively minimizes the mismatch between learned 3D and 2D representations from multiple VLM and VFM image encoders, quantified by differences in the U, T, and G (\cref{fig:contrastive_sim_rel_pointlosses}).
Consequently, the resulting 3D representations significantly outperform those learned via contrastive distillation on zero-shot segmentation tasks. Furthermore, compared to the similarity loss, our relational loss results in 3D representations that consistently improve in-distribution and out-of-distribution few-shot segmentation tasks.
\end{itemize}

\section{Related Work}
\label{sec:related_work}

\subsection{Cross-Modal Knowledge Distillation}
Knowledge Distillation enables a student network to learn a task by mimicking the output of highly-performing teacher networks~\cite{KDsurvey}, achieved by using these output as training targets for the student network~\cite{hinton2015distilling}, or by supervising the student network using the teacher's intermediate representations~\cite{romero2014fitnets, zagoruyko2016paying_attentionKD, tian2019contrastive_CRD}. Distilling representations rather than network predictions enables knowledge transfer across different modalities, without requiring labels. 
In the context of cross-modal distillation, contrastive distillation~\cite{tian2019contrastive_CRD, girdhar2023imagebind} has shown to be the most effective. However, the abundance of self-similarity~\cite{stcl}, coupled with the hardness-aware property of contrastive losses~\cite{wang2021understanding} limits the effectiveness of contrastive losses for AD data. 
On the other hand, similarity losses 
only utilize positive pairs for distillation, resulting in an under-constrained loss, leading to sub-optimal performance on zero-shot tasks~\cite{girdhar2023imagebind}. 
\subsection{3D Representations for Few-Shot Learning}
Few-shot learning refers to learning the underlying pattern in data from only a few training samples~\cite{parnami2022fewlearning}. Distilling 3D representations from 2D self-supervised models~\cite{chen2020_simclr, constrastive_moco, distillation_byol, bardes2022vicreg, caron2021emerging_dino, caron2020_swav} or VFMs~\cite{oquab2023dinov2, radford2021learningclip} has shown to be effective at significantly improving performance on 3D few-shot tasks~\cite{sautier2022slidr, stcl, liu2023seal}. PPKT~\cite{liu2021ppkt} proposes a pixel-based contrastive loss to distill 2D self-supervised representations to point cloud encoders. While effective in indoor settings, PPKT~\cite{liu2021ppkt} underperforms in outdoor settings where point-to-pixel correspondences are sparse~\cite{sautier2022slidr}. Inspired by DetCon~\cite{henaff2021efficient_detcon}, SLidR~\cite{sautier2022slidr} proposes a superpixel-based contrastive loss primarily designed for autonomous driving scenes which constructs region-level contrastive pairs suited for distilling scene-level images~\cite{henaff2021efficient_detcon}. Due to the abundance of self-similarity in AD data~\cite{sautier2022slidr, stcl, liu2023seal}, contrastive distillation leads to sub-optimal 3D representations~\cite{stcl}. 
To address these challenges, ST-SLidR~\cite{stcl} proposes a semantically tolerant contrastive loss 
leading to improved 3D representations for 3D few-shot segmentation tasks. Finally, Seal~\cite{liu2023seal} demonstrates that object priors from VFMs like SAM~\cite{Kirillov_2023_ICCV_sam}, represented as superpixels, can improve the quality of 3D representations for 3D few-shot segmentation tasks. In this paper, we demonstrate that pixel and superpixel-based contrastive distillation applied to AD data, learn 3D representations that significantly differ from the structure of 2D representations. This leads to poor performance on zero-shot tasks and unpredictable performance on few-shot tasks.
\subsection{3D Representations for Zero-Shot Learning}
Contrastive Language-Image Pre-training (CLIP) models are pre-trained on billions of webscale image-text pairs and have shown great success in solving zero-shot image
classification tasks~\cite{radford2021learningclip}. Using frozen CLIP vision encoders, ImageBind~\cite{girdhar2023imagebind} enables zero-shot image classification by distilling image-level representations via a contrastive distillation framework. 
LidarCLIP~\cite{Hess_2024_WACV_LidarCLIP} aligns LiDAR point features to CLIP space, demonstrating effective cross-modal retrieval and image-level zero-shot classification. PointCLIP~\cite{zhang2022pointclip} and PointCLIPv2~\cite{zhu2022pointclipv2} propose a distillation-free approach utilizing CLIP vision and text encoders during the inference stage to solve 3D zero-shot classification tasks. More recently, MaskCLIP~\cite{zhou2022extractMaskCLIP} proposes removing the last attention pooling layer in CLIP vision encoders to enable dense feature extraction for 2D zero-shot segmentation tasks. OpenScene~\cite{peng2023openscene} distills image features from CLIP models fine-tuned on 2D segmentation datasets~\cite{lseg, ghiasi2022scaling_openseg} to point cloud encoders. As demonstrated by ConceptFusion~\cite{conceptfusion} and LERF~\cite{kerr2023lerf}, due to fine-tuning on closed-set 2D segmentation labels, these CLIP models have poor open-set capabilities. During inference, OpenScene fuses image and point features to solve 3D zero-shot segmentation. In this work, we do not assume access to finetuned CLIP models or access to image data during inference time. CLIP2Scene~\cite{chen2023clip2scene} assumes knowledge of class names of objects in the pre-training dataset, which leads to a contrastive loss with fewer false negatives. Requiring class names is problematic as it assumes a dataset only consists of a predefined set of classes, preventing the transfer of features associated with undefined classes and limiting open-set capabilities~\cite{puy2023revisit}. 

\section{Methodology}
\label{sec:methodology}

\subsection{Preliminaries}
Our objective is to generate useful 3D representations by learning a point cloud encoder, $f: \mathbb{R}^{N \times (3 + L)} \to \mathbb{R}^{N \times C}$, through distilling 2D representations from VFMs or VLMs (e.g., CLIP~\cite{radford2021learningclip}, DINOv2~\cite{oquab2023dinov2}). Using camera-to-LiDAR calibration matrices, we create a set of positive pairs of point, $\mathbf{K_p} \in \mathbb{R}^{N \times C}$, and pixel, $\mathbf{Q_p} \in \mathbb{R}^{N \times C}$, features, with the latter being generated from the pre-trained image encoder of the foundation model, $g: \mathbb{R}^{N \times 3} \to \mathbb{R}^{N \times C}$.
While distillation methods like PPKT~\cite{liu2021ppkt} implement distillation losses by creating pixel-point positive pairs, techniques such as SLidR~\cite{sautier2022slidr} optimize pair formation by harnessing boundary information from superpixels. SLidR, with $M$ superpixels per image, employs average pooling to group points, $\mathbf{K_p}$, and pixels, $\mathbf{Q_p}$, into superpoint, $\mathbf{K_{sp}} \in \mathbb{R}^{M \times C}$, and superpixel, $\mathbf{Q_{sp}} \in \mathbb{R}^{M \times C}$, features, improving feature correspondence. Superpixels are derived from unsupervised techniques like SLIC~\cite{slic}, or foundation models such as SAM~\cite{liu2023seal, Kirillov_2023_ICCV_sam}.
One of the most effective cross-modal distillation losses is the contrastive loss:
\begin{align}
    \mathcal{L}_{con}\left(\mathbf{K}, \mathbf{Q}\right)=- \frac{1}{B} {\sum_{i=1}^{B} \log \left[\frac{e^{\left(\left<\mathbf{k}_i,\mathbf{q}_i\right>/\tau\right)}}{\sum_{j \neq i} e^{\left(\left<\mathbf{k}_i,\mathbf{q}_j\right>/\tau\right)}+e^{\left(\left<\mathbf{k}_i,\mathbf{q}_i\right>/\tau\right)}}\right]}
    \label{eq:loss_contrastive}
\end{align}
where $\mathbf{K}$ and $\mathbf{Q}$ can either be point/pixel (i.e., $\mathbf{K_p}$, $\mathbf{Q_p}$) or superpoint/superpixel (i.e., $\mathbf{K_{sp}}$, $\mathbf{Q_{sp}}$)  feature vectors, $B$ is the number of positive pairs in a mini-batch, $\tau$ is the temperature, and $\left<\mathbf{k}_i,\mathbf{q}_j\right>$ is the dot product between the $\ell_2$-normalized features. The contrastive loss distills information from pre-trained image encoders by pulling the point cloud features, $K$, towards their corresponding (positive) image feature, $Q$, in representation space, simultaneously pushing them away from all the other (negative) image features. The temperature, $\tau$, controls the strength of this push/pull mechanism by modifying the gradient's scale from the negative samples~\cite{wang2021understanding}. However, a significant limitation of this approach is the potential degradation in the learned point cloud representation's quality due to false negative samples~\cite{stcl}. These are image features incorrectly chosen as negative, despite belonging to the same semantic class as the positive point cloud feature, leading to opposing distillation signals. 

To avoid both relying on negative examples while performing distillation, and tuning the temperature parameter, we can use the cosine similarity:
\begin{align}
    \mathcal{L}_{sim}\left(\mathbf{K}, \mathbf{Q}\right) = \frac{1}{B} \sum_{i=1}^{B} \left(1.0 - \left<\mathbf{k}_i,\mathbf{q}_i\right> \right)
    \label{eq:loss_similarity}
\end{align}
The similarity loss does not rely on negative samples and thus has a much simpler mode of action. It focuses solely on drawing each point cloud feature, $\mathbf{k}_i$, nearer to its corresponding image feature, $\mathbf{q}_i$, effectively maximizing their dot product. 

\subsection{Quantifying the Quality of Distilled Representations}
We investigate the quality of the distilled 3D representations as a function of the distillation loss used during training. We hypothesize that if a distilled 3D representation space closely captures the structure of the source 2D representation space, our resulting 3D encoder would: 1) possess the enhanced representation capability of the source vision/vision-language foundation model and 2) generate 3D representations that are well-aligned with the text representations of the vision-language model, enabling zero-shot downstream tasks. 
To assess the structural similarity between two representation spaces, we utilize uniformity and tolerance, two metrics previously proposed by Wang et al.~\cite{wang2021understanding}, for evaluating the quality of a representation space. 
Uniformity measures the distribution of $\ell_2$-normalized features on a hyper-sphere. Authors in~\cite{wang2020understanding_uniformity_alignment} have demonstrated that a high uniformity is key for high-quality representations as it quantifies that the trained encoder has successfully utilized a substantial part of the available feature space. Uniformity is formulated using a Gaussian potential function as:
\begin{align}
\text U(f(.))=-\log \underset{x, y \sim p_{\text {data }}}{\mathbb{E}}\left[e^{-t|| f(x)-f(y) \|_2^2}\right] 
\label{eq:uniformity}
\end{align}
where $f(.)$ is the point encoder and $x,y$ are two samples from the data-distribution $p_{\text {data}}$. Here, $x,y$ are point features for 3D point encoders. Similarly, we compute the uniformity of $U(g(.))$ for the image encoder by setting $x,y$ as pixel features.

On the other hand, tolerance measures the semantic clustering for the $\ell_2$-normalized output representations from a given encoder,computed as:
\begin{align}
\label{eq:tolerance}
T(f(.))=\underset{x, y \sim p_{\text {data }}}{\mathbb{E}}\left[\left(f(x)^T f(y)\right) \cdot I_{l(x)=l(y)}\right]
\end{align}
where $l(x)$ represents the class label of point $x$. $I_{l(x)=l(y)}$
is an indicator function, having the value of 1 for $l(x) = l(y)$ and the value of 0 for $l(x) \neq l(y)$. A higher tolerance indicates that the features of all points belonging to a certain class are better clustered together on a unit sphere. Similar to uniformity, we can compute tolerance for the image encoder as $T(g(.))$, by propagating 3D point-wise labels to 2D image pixels.

Furthermore, to directly compare the alignment of the two representation spaces from different modalities, we also use the modality gap as proposed in~\cite{liang2022mind}:
\begin{equation}
\label{eq:modality_gap}
G(f(.), g(.))=\underset{x \sim p_{\text {2D}}}{\mathbb{E}} \left[f(x)\right]-\underset{y \sim p_{\text {3D}}}{\mathbb{E}}  \left[g(y)\right]
\end{equation}
The modality gap measures the difference between the mean of the representation of the point features and their corresponding pixel features. 
\begin{figure}[t]
  \centering
  \includegraphics[width=0.7\textwidth]{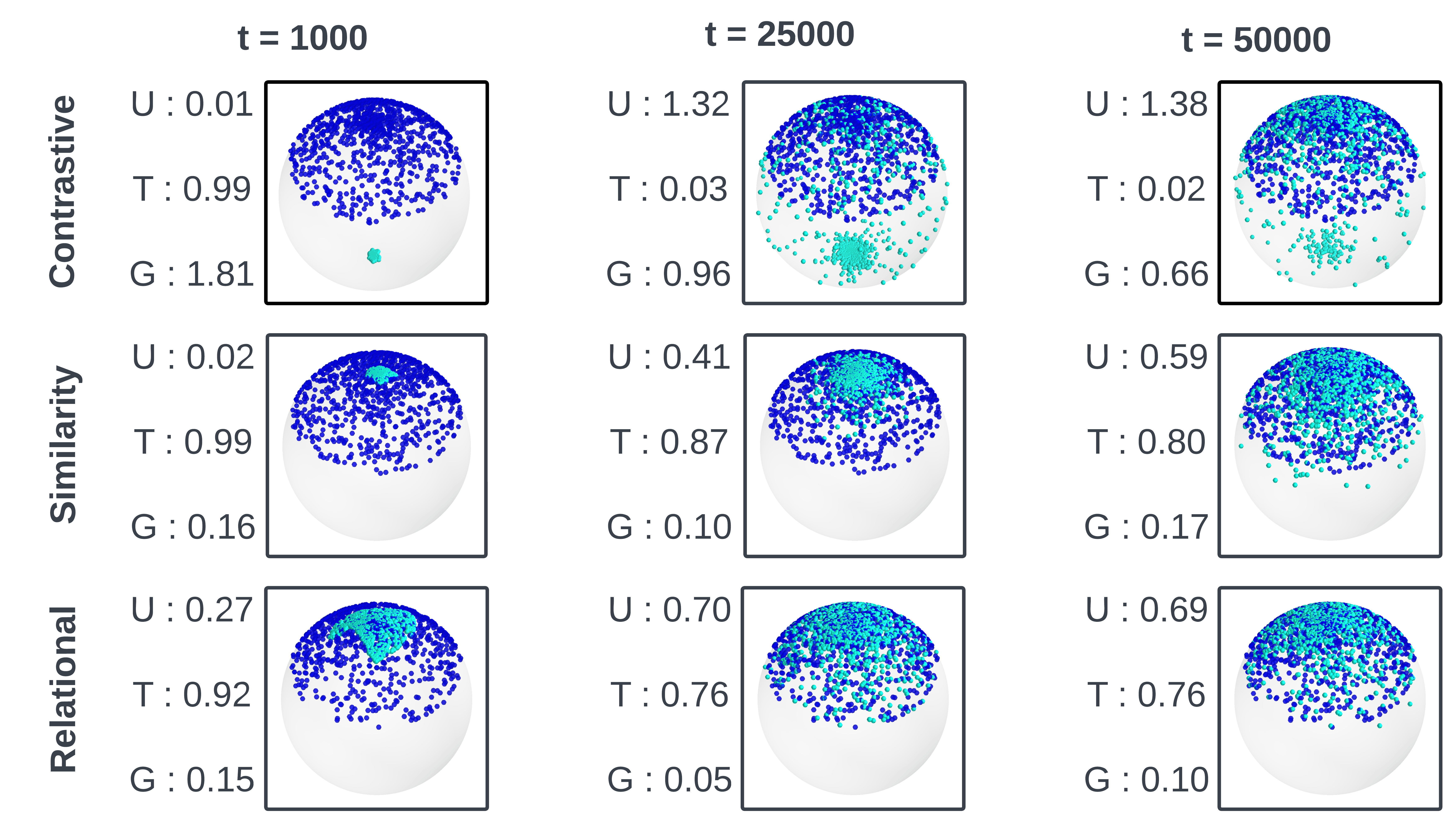}
  \caption{\textbf{Blue:} The source representation space, with a uniformity and tolerance of U=0.89 and T=0.66, respectively. \textbf{Cyan:} The predicted representation space. Here, $t$ denotes the number of training iterations.}
  \label{fig:toy_example}
  \vspace{-15pt}
\end{figure}
We conclude by noting that unlike G, U and T are properties of a single encoder; we show the difference in representations by comparing the values of $U(f(.))$ and $T(f(.))$ to $U(g(.))$ and $ T(g(.))$, respectively. Closer values indicate our point cloud encoder well-approximates the representation space of the source image encoder, which we show in Section~\ref{sec:experiments} to be beneficial for downstream tasks.

\subsection{The Representations of Common Distillation Losses}

Inspired by~\cite{shi2023towards}, we explore the structure of the learned representation space using contrastive loss and similarity loss through a toy example. We start with $1000$ uniformly distributed points over a 3D unit sphere, representing point features before the distillation phase. Using a 2-layer MLP, we learn to align each input point with its corresponding pixel feature from the source representation space, defined by U and T levels of $0.89$ and $0.66$, respectively. We use the Adam optimizer with a learning rate of $10^{-4}$ and train our model for $50,000$ iterations (refer to Appendix~\ref{app:toy_example} for detailed analysis). 

Figure~\ref{fig:toy_example} shows a visualization of the source representation space (blue) and the predicted representation space (cyan) at various stages of training. The top row highlights a significant difference: the contrastive losses generate a representation space with a high U of $1.38$, but with a very low T of $0.02$, differing significantly from the source space's U ($0.89$) and T ($0.66$). This is attributed to self-similarity~\cite{stcl}, a phenomenon that occurs when points from the source representation space are close to one another, as in our toy example. Although the contrastive loss moves a particular prediction to its corresponding point in the source, its hardness-awareness property~\cite{wang2021understanding} pushes all other predictions, particularly close ones, away from that point. Combined with self-similarity, this yields a uniform predicted space with a low T, and a significant modality gap. As demonstrated in Section~\ref{sec:experiments}, this discrepancy hinders zero-shot task performance due to inadequate cross-modal alignment.



On the other hand, learning with similarity loss leads to representations with lower U ($0.59$) and higher T ($0.80$) compared to the source. This stems from the tendency of similarity loss to form compact clusters in the predicted space, as shown with autonomous driving data in Figure~\ref{fig:contrastive_sim_rel_pointlosses} and the toy example in Figure~\ref{fig:toy_example}. Notably, early training (t=1000) exhibits noticeable clustering that disperses with additional epochs. Additionally, the non-uniqueness of the dot product between the learnable point vector, $\mathbf{k}_i$, and the fixed pixel vector, $\mathbf{q}_i$, within the framework of similarity loss (Eq.~\ref{eq:loss_similarity}) results in a significant increase in the number of learnable vectors that can achieve the same cosine similarity with the fixed vector, $\mathbf{q}_i$. This growth can result in slow convergence and suboptimal predicted representation spaces. Combined with neural network optimization complexities, these issues lead to the observed disparity in uniformity and tolerance from the source representation space in distillation tasks using similarity loss.

\subsection{Relational Loss}
Inspired by distilling relations in model compression~\cite{park2019relational_RKD}, this section presents two relational losses that impose structural constraints on the 3D representation space. 
This modification disrupts the symmetry inherent in the similarity loss. It drives the network to select solutions that, while possessing equivalent similarity loss values to alternatives, yield a 3D representation space that more accurately mirrors the structure of the image representation space.

\noindent \textbf{Cross-modal Relational Loss} We propose imposing a constraint on the structure of the learned 3D representation space to ensure that the similarities between a given predicted point feature, $\mathbf{k}_i$, and all source pixel features align with the similarities between its corresponding pixel feature, $\mathbf{q}_i$, and all other source pixel features in the same batch. The cross-modal relation loss is defined as: 
\begin{align}
\label{eqn:cross-modal}
\mathcal{L}_{cross}\left(\mathbf{K}, \mathbf{Q}\right) &= \frac{1}{N} \sum_{i=1}^{N} \mathbf{C}_{ii} + \frac{1}{N^2 - N} \sum_{i=1}^{N}  \sum_{\substack{j = 1 \\ i \neq j}}^{N} \mathbf{C}_{ij}
\nonumber \\
&= \underbrace{\frac{1}{N} \sum_{i=1}^{N} \left(1.0 - \left<\mathbf{k}_i, \mathbf{q}_i\right> \right)}_{\mathcal{L}_{sim}\left(\mathbf{K}, \mathbf{Q}\right)} + \frac{1}{N^2 - N} \sum_{i=1}^{N}  \sum_{\substack{j = 1 \\ i \neq j}}^{N}  |\left<\mathbf{k}_i, \mathbf{q}_j\right> -  \left<\mathbf{q}_i, \mathbf{q}_j\right>|
\end{align}
where $\mathbf{C} = | \mathbf{K}\mathbf{Q}^T - \mathbf{Q}\mathbf{Q}^T|$ represents the matrix of differences, capturing the discrepancy between the predicted point-to-pixel and the source pixel-to-pixel similarities. First, the matrix's diagonal components, $\mathbf{C}_{ii}$, are directly linked to the similarity loss, underscoring the necessity of aligning each point feature with its corresponding source pixel feature. The second term of Equation~\ref{eqn:cross-modal} facilitates learning a point feature, $\mathbf{k}_i$, that is not only close to its corresponding source pixel feature, $\mathbf{q}_i$, but is also consistent with $\mathbf{q}_i$'s similarities to other pixel features, $\left<\mathbf{q}_i, \mathbf{q}_j\right>$, where $j\neq i$. These cross-modal constraints foster learning point features that maintain the relational structure within the image representation space. 

\noindent \textbf{Intra-modal Relational Loss} Another strategy to ensure structural similarity between the 3D and 2D representation spaces involves directly penalizing differences in their relational graphs. A relational graph of a representation space can be understood as a graph with the nodes representing point or pixel features, and edges indicating the similarity between node features. We represent the relational graph of the point features and the pixel features using $\mathbf{K}\mathbf{K}^T$ and $\mathbf{Q}\mathbf{Q}^T$, respectively. Here, $\left<\mathbf{k}_i, \mathbf{k}_j\right>$ represents the similarity between the $i^{th}$ and the $j^{th}$ predicted point feature. We denote the discrepancy of the relational graphs as $\mathbf{U} = | \mathbf{K}\mathbf{K}^T - \mathbf{Q}\mathbf{Q}^T|$. Since this matrix is symmetric, and its diagonal elements, $ | \left<\mathbf{k}_i, \mathbf{k}_i\right> - \left<\mathbf{q}_i, \mathbf{q}_i\right>|$, degenerate to 0, the intra-modal relational loss is expressed as:
\begin{align}
\label{eqn:unimodal}
\mathcal{L}_{intra}\left(\mathbf{K}, \mathbf{Q}\right) &=\frac{2}{N^2 - N} \sum_{i=1}^{N-1} \sum_{j=i+1}^{N} \mathbf{U}_{ij}
\end{align}
The consequence of this loss on the structure of the point representation space is simple; the similarity between a predicted point feature and other predicted point features in a batch should match the similarity between its corresponding source pixel feature and other source pixel features in the same batch. Our proposed relational loss is a combination of the two losses:
\begin{align}
\mathcal{L}_{rel}\left(\mathbf{K}, \mathbf{Q}\right) &= \mathcal{L}_{intra}\left(\mathbf{K}, \mathbf{Q}\right) + \mathcal{L}_{cross}\left(\mathbf{K}, \mathbf{Q}\right)
\end{align}

Unlike the similarity loss, the proposed loss ensures learned point cloud features align with image features in a structured manner, preserving the data's inherent relationships. Figure~\ref{fig:toy_example} reveals two key advantages of relational loss over similarity loss: closer alignment of the predicted space's U and T with those of the source space, alongside a reduced G between the two spaces. Additionally, the relational loss achieves faster convergence, as the visual appearance of the predicted representation space resulting from learning with the relational loss remains consistent from the mid-point of training (t=25000) to its conclusion (t=50000). In Section~\ref{sec:experiments}, we further illustrate how these attributes of the relational loss contribute to enhanced performance in downstream tasks.

\section{Experiments} 
\label{sec:experiments}
\subsection{Pre-training}
\noindent\textbf{Backbones} We focus on distilling 2D representations from two vision foundation models, CLIP~\cite{radford2021learningclip}, pretrained on the WIT dataset containing 400 million image-text pairs,  and DINOv2~\cite{oquab2023dinov2}, pretrained on the LVD-142 dataset containing 142 million images. For CLIP encoders, we experiment with different pre-trained architectures available from OpenAI~\cite{radford2021learningclip}. 
For the 3D backbone, we use a randomly initialized Minkowski U-Net~\cite{sautier2022slidr}. For details on the design of the 2D and 3D projection layers, refer to~\cref{app:proj_layers}.
\\
\noindent\textbf{Dataset} In the pre-training phase we use the nuScenes dataset~\cite{caesar2020nuscenes}, consisting of 700 scenes. In line with previous works~\cite{sautier2022slidr, stcl, liu2023seal}, these scenes are divided into two subsets: 600 scenes for pre-training and 100 scenes for the determination of the best hyper-parameters. Throughout this pre-training process, we exclusively employ keyframes from the 600 scenes to train all models. For pre-training hyperparameters and data-augmentation details, refer to~\cref{app:training}.

\subsection{Evaluation}
To study the effect of the distillation framework on the structure of the 3D representations, we evaluate the average U, T, and G between the distilled 3D and source 2D representations. For both pixel-based and superpixel-based losses, U, T, and G are evaluated using point or pixel features. To assess the relationship between the performance on downstream tasks and the difference in the structure of the 2D and the 3D representation spaces, we use the in-distribution 3D few-shot segmentation task, where we learn a classifier by finetuning the 3D representations on 1\% of the labels of the nuScenes training set. Similar to~\cite{sautier2022slidr, stcl}, we also evaluate the utility of the distilled representations in the out-of-distribution setting by fine-tuning on 1\% of the SemanticKITTI~\cite{behley2019iccv_semkitti} dataset. The nuScenes and SemanticKITTI datasets contain 16 and 19 classes, respectively. We present our results using the official validation sets for these datasets. 

For DINOv2, we follow~\cite{sautier2022slidr, stcl} and evaluate the Linear Probing performance on nuScenes by freezing the backbone of the point cloud encoder and training a linear classifier on 100\% of the labels. 
For CLIP models, we evaluate the 3D zero-shot segmentation tasks which can be performed with language prompts. To enable openset scene understanding, during pretraining, we assume we have no access to nuScenes class labels. Similar to \cite{zhou2022extractMaskCLIP}, we apply prompt engineering during inference using 85 hand-crafted prompts for each class label, and then use the CLIP text encoder to compute an average text embedding for each class. Each point is then assigned the label that corresponds to the highest cosine similarity, computed between the point features and the CLIP text embeddings. 
\begin{table}[t]
\centering
\caption{Evaluation of 3D representations from CLIP image encoder with source U of 1.54, and source T of 0.73. Best results are bolded and second best are underlined.}
\setlength{\tabcolsep}{0.75em}
\scalebox{0.75}
{
\begin{tabular}{clcccccc}
\toprule
\multirow{3}{*}{\begin{tabular}[c]{@{}c@{}}2D \\ Encoder\end{tabular}} & \multicolumn{1}{c}{\multirow{3}{*}{\begin{tabular}[c]{@{}c@{}}Distillation \\ Loss\end{tabular}}} & \multicolumn{1}{c}{\multirow{3}{*}{Uniformity}} & \multicolumn{1}{c}{\multirow{3}{*}{Tolerance}} & \multicolumn{1}{c}{\multirow{3}{*}{\begin{tabular}[c]{@{}c@{}}Modality \\ Gap\end{tabular}}} & \multicolumn{2}{c}{nuScenes}                                                                            & KITTI                                                                     \\ \cline{6-8} 
                                                                       & \multicolumn{1}{c}{}                      & \multicolumn{1}{c}{}                            & \multicolumn{1}{c}{}                           & \multicolumn{1}{c}{}                                                                         & \multirow{2}{*}{Zero-shot} & \multirow{2}{*}{\begin{tabular}[c]{@{}c@{}}Finetuning\\  1\%\end{tabular}} & \multirow{2}{*}{\begin{tabular}[c]{@{}c@{}}Finetuning\\ 1\%\end{tabular}} \\
                                                                       & \multicolumn{1}{c}{}                      & \multicolumn{1}{c}{}                            & \multicolumn{1}{c}{}                           & \multicolumn{1}{c}{}                                                                         &                            &                                                                            &                                                                           \\ \hline
\multicolumn{1}{c|} {\multirow{7}{*}{CLIP~\cite{radford2021learningclip}}}                                                  & PPKT~\cite{liu2021ppkt}                                      & 3.5210                                           & 0.5217                                         & 0.00158                                                                                     & 14.53                      & \underline{45.31}                                                                      & 45.77                                                                     \\
                                                                       \multicolumn{1}{c|}{} &  $\text{Sim}_{pl}$                                    & 0.5519                                          & 0.9477                                         & 0.00045                                                                                      & 20.84                      & 44.39                                                                       & 45.60                                                                      \\
                                                                       \multicolumn{1}{c|}{} &  $\text{Rel}_{pl}$ (ours)                                    & 0.9089                                          & 0.9145                                         & 0.00033                                                                                    & \textbf{23.53}                      & \textbf{45.67}                                                                      & \textbf{46.06}                                                                     \\ \cline{2-8} 
                                                                       \multicolumn{1}{c|}{} & SLidR~\cite{sautier2022slidr}                                     & 3.5090                                           & 0.4472                                         & 0.00153                                                                                      & 16.82                      & 46.76                                                                      & 46.53                                                                     \\
                                                                       \multicolumn{1}{c|}{} &  ST-SLidR~\cite{stcl}                                  & 3.5000                                             & 0.4544                                         & 0.00150                                                                                       & 18.54                      & \underline{47.13}                                                                      & \underline{46.84}                                                                     \\
                                                                       \multicolumn{1}{c|}{} & $\text{Sim}_{spl}$                                    & 0.5919                                          & 0.9333                                         & 0.00040                                                                                       & \underline{23.93}                      & 45.63                                                                      & 46.42                                                                     \\
                                                                       \multicolumn{1}{c|}{} & $\text{Rel}_{spl}$ (ours)                                    & 1.0990                                           & 0.8700                                           & 0.00019                                                                                      & \textbf{26.03}                      & \textbf{47.22}                                                                      & \textbf{47.52}                                                            \\
                                                                       \bottomrule
\end{tabular}}
\label{tab:clip_baseline}
\end{table}
\begin{table}[t]
\centering
\caption{Evaluation of 3D representations from DINOv2 image encoder with source U of 2.8, and source T of 0.51. Best results are bolded and second best are underlined.}
\setlength{\tabcolsep}{0.7em}
\scalebox{0.73}{
\begin{tabular}{clcccccc}
\toprule
\multirow{3}{*}{\begin{tabular}[c]{@{}c@{}}2D \\ Encoder\end{tabular}} & \multicolumn{1}{c}{\multirow{3}{*}{\begin{tabular}[c]{@{}c@{}}Distillation \\ Loss\end{tabular}}} & \multicolumn{1}{c}{\multirow{3}{*}{Uniformity}} & \multicolumn{1}{c}{\multirow{3}{*}{Tolerance}} & \multicolumn{1}{c}{\multirow{3}{*}{\begin{tabular}[c]{@{}c@{}}Modality \\ Gap\end{tabular}}} & \multicolumn{2}{c}{nuScenes}                                                                                                                                & KITTI                                                                     \\ \cline{6-8} 
                                                                       & \multicolumn{1}{c}{}                      & \multicolumn{1}{c}{}                            & \multicolumn{1}{c}{}                           & \multicolumn{1}{c}{}                                                                         & \multirow{2}{*}{\begin{tabular}[c]{@{}c@{}}Lin. Probing \\ 100\%\end{tabular}} & \multirow{2}{*}{\begin{tabular}[c]{@{}c@{}}Finetuning \\ 1\%\end{tabular}} & \multirow{2}{*}{\begin{tabular}[c]{@{}c@{}}Finetuning\\ 1\%\end{tabular}} \\
                                                                       & \multicolumn{1}{c}{}                      & \multicolumn{1}{c}{}                            & \multicolumn{1}{c}{}                           & \multicolumn{1}{c}{}                                                                         &                                                                                &                                                                            &                                                                           \\ \hline
\multicolumn{1}{c|}{\multirow{7}{*}{DINOv2~\cite{oquab2023dinov2}}}                           & PPKT~\cite{liu2021ppkt}                                      & 3.625                                           & 0.4451                                         & 0.00060                                                                                       & 48.50                                                                          & 43.30                                                                      & 43.62                                                                     \\
\multicolumn{1}{c|}{}                                                  & $\text{Sim}_{pl}$                                 & 2.176                                           & 0.6926                                         & 0.00030                                                                                       & \textbf{49.72}                                                                          & \underline{45.85}                                                                      & \underline{48.32}                                                                     \\
\multicolumn{1}{c|}{}                                                  & $\text{Rel}_{pl}$ (ours)                                 & 2.393                                           & 0.6588                                         & 0.00020                                                                                       & 49.20                                                                          & \textbf{46.90}                                                                      & \textbf{49.00}                                                                     \\ \cline{2-8} 
\multicolumn{1}{c|}{}                                                  & SLidR~\cite{sautier2022slidr}                                     & 3.655                                           & 0.3739                                         & 0.00044                                                                                      & 49.60                                                                          & 44.81                                                                      & 42.23                                                                     \\
\multicolumn{1}{c|}{}                                                  & ST-SLidR~\cite{stcl}                                  & 3.589                                           & 0.4326                                         & 0.00042                                                                                      & \textbf{53.00}                                                                          & 47.11                                                                      & 45.61                                                                     \\
\multicolumn{1}{c|}{}                                                  & $\text{Sim}_{spl}$                                    & 2.286                                           & 0.6522                                         & 0.00030                                                                                       & 52.30                                                                          & \underline{47.23}                                                                      & \underline{49.01}                                                                     \\
\multicolumn{1}{c|}{}                                                  & $\text{Rel}_{spl}$ (ours)                                    & 2.504                                           & 0.6312                                         & 0.00023                                                                                      & \underline{52.92}                                                                          & \textbf{48.42}                                                                      & \textbf{49.10}                                           \\ \bottomrule                         
\end{tabular}}
\label{tab:dino_baseline}
\end{table}
\subsection{Results}
\noindent\textbf{Baselines} 
In~\cref{tab:clip_baseline} and ~\cref{tab:dino_baseline}, we present the results of distilling CLIP and DINOv2 models using our proposed relational loss. We compare against multiple state-of-the-art contrastive losses including PPKT~\cite{liu2021ppkt}, SLidR~\cite{sautier2022slidr}, and ST-SLidR~\cite{stcl}. For superpixel-based losses, masks are generated using SAM~\cite{Kirillov_2023_ICCV_sam}. As an additional baseline, we report results for both the pixel-based and superpixel-based similarity loss, which we denote as $\text{Sim}_{pl}$ and $\text{Sim}_{spl}$, respectively. We report the average over $3$ runs for all metrics. 

\noindent\textbf{Uniformity, tolerance, and closing the modality gap}
Looking at distillation losses from the 2D CLIP encoder in~\cref{tab:clip_baseline}, we first observe that U of 3D representations distilled using contrastive loss (i.e., PPKT, SLidR and ST-SLidR) is higher than the source U of 1.54, while U of similarity loss (i.e., $\text{Sim}_{pl}$, $\text{Sim}_{spl}$) is lower than the U of the source. This is also observed when distilling from DINOv2. The relational constraints (i.e., $\text{Rel}_{pl}$, $\text{Rel}_{spl}$) effectively close the gap between 3D representations distilled by similarity loss (i.e., $\text{Sim}_{pl}$, $\text{Sim}_{spl}$) and 2D representations. For instance, when the pixel-based loss $\text{Rel}_{pl}$ is applied to CLIP, relational constraints close the gap in U resulting in a 3D representation with a U of 0.9089, instead of 0.5519, thus closer to the U of the source (1.54). Similarly, for the superpixel-based loss, $\text{Rel}_{spl}$, relational constraints close the gap in U resulting in a 3D representation with a U of 1.099, instead of 0.5919, thus closer to the U of the source (1.54). In addition, relational losses result in semantic clusters that are closer to the T of the source. By closing the gap in U and T, relational constraints result in the lowest G for pixel-based and superpixel-based losses. These results align well with the ones presented on the toy example in Section~\ref{sec:methodology}, which further supports the validity of our analysis.
\begin{table}[t]
\centering
\caption{Performance for different CLIP backbones using pixel-based and superpixel-based relational loss. Best results are bolded and second best are underlined.}
\setlength{\tabcolsep}{0.7em}\scalebox{0.75}{
\begin{tabular}{clcccccc}
\toprule
\multirow{3}{*}{\begin{tabular}[c]{@{}c@{}}2D \\ Encoder\end{tabular}} & \multicolumn{1}{c}{\multirow{3}{*}{\begin{tabular}[c]{@{}c@{}}Distillation \\ Loss\end{tabular}}} & \multicolumn{1}{c}{\multirow{3}{*}{Uniformity}} & \multicolumn{1}{c}{\multirow{3}{*}{Tolerance}} & \multicolumn{1}{c}{\multirow{3}{*}{\begin{tabular}[c]{@{}c@{}}Modality \\ Gap\end{tabular}}} & \multicolumn{2}{c}{nuScenes}                                                                                                                                  & \multicolumn{1}{c}{KITTI}                                                             \\ \cline{6-8} 
                                                                       & \multicolumn{1}{c}{}                      & \multicolumn{1}{c}{}                            & \multicolumn{1}{c}{}                           & \multicolumn{1}{c}{}                                                                         & \multicolumn{1}{c}{\multirow{2}{*}{Zero-shot}} & \multicolumn{1}{c}{\multirow{2}{*}{\begin{tabular}[c]{@{}c@{}}Finetuning\\  1\%\end{tabular}}} & \multicolumn{1}{c}{\multirow{2}{*}{\begin{tabular}[c]{@{}c@{}}Finetuning\\ 1\%\end{tabular}}} \\
                                                                       & \multicolumn{1}{c}{}                      & \multicolumn{1}{c}{}                            & \multicolumn{1}{c}{}                           & \multicolumn{1}{c}{}                                                                         & \multicolumn{1}{c}{}                    & \multicolumn{1}{c}{}                                                                   & \multicolumn{1}{c}{}                                                                  \\ \hline
\multicolumn{1}{c|}{\multirow{6}{*}{ViT-B32}}                          & PPKT                                      & 3.5110                                           & 0.5203                                         & 0.00020                                                                                       & 14.21                                   & 43.29                                                                                  & \textbf{45.89}                                                                                 \\
\multicolumn{1}{c|}{}                                                  & Sim$_{pl}$                                    & 0.5804                                          & 0.9389                                         & 0.00010                                                                                       & 22.84                                   & \underline{44.22}                                                                                  & 44.30                                                                                 \\
\multicolumn{1}{c|}{}                                                  & Rel$_{pl}$ (ours)                                & 0.9605                                          & 0.8989                                         & 0.00010                                                                              & \textbf{23.85}                          & \textbf{44.67}                                                                         & \underline{45.62}                                                                        \\ \cline{2-8} 
\multicolumn{1}{c|}{}                                                  & SLidR                                     & 3.5130                                           & 0.4706                                         & 0.00030                                                                                       & 15.55                                   & 43.38                                                                                  & 43.82                                                                                 \\
\multicolumn{1}{c|}{}                                                  & Sim$_{spl}$                                & 0.6009                                          & 0.9323                                         & 0.00011                                                                                      & \underline{25.46}                        & \underline{44.83}                                                                      & \underline{45.69}                                                                        \\
\multicolumn{1}{c|}{}                                                  & Rel$_{spl}$ (ours)                                & 1.2020                                           & 0.8648                                         & 0.00007                                                                             & \textbf{25.66}                          & \textbf{45.60}                                                                         & \textbf{47.26}                                                                           \\ \hline
\multicolumn{1}{c|}{\multirow{6}{*}{ViT-B16}}                          & PPKT                                      & 3.5210                                           & 0.5217                                         & 0.00158                                                                                     & 14.53                                   & \underline{45.31}                                                                         & \underline{45.77}                                                                       \\
\multicolumn{1}{c|}{}                                                  & Sim$_{pl}$                                    & 0.5519                                          & 0.9477                                         & 0.00045                                                                                   & \underline{20.84}                       & 44.39                                                                                  & 45.59                                                                                 \\
\multicolumn{1}{c|}{}                                                  & Rel$_{pl}$ (ours)                                & 0.9089                                          & 0.9145                                         & 0.00033                                                                                    & \textbf{23.53}                          & \textbf{45.67}                                                                         & \textbf{46.06}                                                                         \\ \cline{2-8} 
\multicolumn{1}{c|}{}                                                  & SLidR                                     & 3.5090                                           & 0.4472                                         & 0.00153                                                                                      & 16.82                                   & \underline{46.76}                                                                         & \underline{46.53}                                                                       \\
\multicolumn{1}{c|}{}                                                  & Sim$_{spl}$                                & 0.5919                                          & 0.9333                                         & 0.00040                                                                                   & \underline{23.93}                       & 45.63                                                                                  & 46.42                                                                                 \\
\multicolumn{1}{c|}{}                                                  & Rel$_{spl}$ (ours)                                & 1.0990                                           & 0.8700                                           & 0.00019                                                                                     & \textbf{26.03}                          & \textbf{47.22}                                                                         & \textbf{47.52}                                                                          \\ \hline
\multicolumn{1}{c|}{\multirow{6}{*}{ViT-L14}}                          & PPKT                                      & 3.5020                                           & 0.5203                                         & 0.00089                                                                                      & 16.28                                   & 44.69                                                                                  & \textbf{46.69}                                                                         \\
\multicolumn{1}{c|}{}                                                  & Sim$_{pl}$                                & 0.6909                                          & 0.9257                                         & 0.00016                                                                                  & \textbf{27.91}                          & \underline{44.87}                                                                      & 44.80                                                                                 \\
\multicolumn{1}{c|}{}                                                  & Rel$_{pl}$ (ours)                                & 0.8775                                          & 0.9027                                         & 0.00013                                                                                    & \underline{27.74}                       & \textbf{45.92}                                                                         & \underline{45.86}                                                                       \\ \cline{2-8} 
\multicolumn{1}{c|}{}                                                  & SLidR                                     & 3.4800                                            & 0.4403                                         & 0.00065                                                                                      & 18.31                                   & \textbf{47.34}                                                                         & \textbf{46.86}                                                                         \\
\multicolumn{1}{c|}{}                                                  & Sim$_{spl}$                                & 0.7311                                          & 0.9102                                         & 0.00020                                                                                      & \textbf{30.77}                          & 45.76                                                                                  & 45.29                                                                                 \\
\multicolumn{1}{c|}{}                                                  & Rel$_{spl}$ (ours)                                & 0.9019                                          & 0.8873                                        & 0.00017                                                                                   & \underline{30.11}                        & \underline{47.07}                                                                      & \underline{46.81}                                                                       \\
\bottomrule
\end{tabular}}
\label{tab:clip_arch}

\end{table}

\noindent\textbf{Zero-shot performance} Looking at the utility of the 3D representations for 3D zero-shot segmentation in~\cref{tab:clip_baseline}, we observe that the contrastive loss (i.e., PPKT, SLidR and ST-SLidR) has significantly worse 3D zero-shot mean IOU when compared to the similarity loss. This is contrary to the conclusion reached by ImageBind~\cite{girdhar2023imagebind}, which observed that the contrastive loss achieves 6\% improvement on 2D zero-shot tasks compared to similarity losses. We hypothesize that the abundance of self-similarity in AD datasets, coupled with the hardness-aware property of contrastive loss~\cite{stcl}, leads to pushing away semantically similar point and pixel features. This, in turn, results in poor alignment between point features and CLIP image features. Notably, we observe that ST-SLidR improves zero-shot performance compared to SLidR (18.54\% vs 16.82\%) as it utilizes the superpixel feature similarities to exclude a portion of the false negative samples from the pool of negative samples. Without resorting to negative samples, relational constraints minimize the 2D-to-3D structural gap compared to similarity loss, leading to improvements in zero-shot performance from 20.84\% to 23.53\% for pixel-based losses, and from 23.93\% to 26.03\% for superpixel-based losses. 

In~\cref{tab:zero_shot_performance}, we compare pixel and superpixel-based relational losses to SOTA methods. All methods distill from CLIP~\cite{radford2021learningclip}, with the performance on nuScenes provided by~\cite{chen2023towards}. Methods are further categorized based on the prior knowledge required during the distillation phase. Some methods require class names defined for a dataset~\cite{chen2023clip2scene, chen2023towards}, while others utilize SAM to refine CLIP predictions~\cite{chen2023towards}. Requiring class names assumes a dataset only consists of a predefined set of classes, preventing the transfer of features associated with undefined classes, thus limiting openset capabilities~\cite{puy2023revisit}. Without using class information, we observe in~\cref{tab:zero_shot_performance} that $\text{Rel}_{pl}$ performs the best compared to other pixel-based losses, while our superpixel-based relational loss, $\text{Rel}_{spl}$, is within 1\% from state-of-the-art~\cite{chen2023towards}.

\noindent\textbf{Few-shot performance} We evaluate the effectiveness of 3D representations in both in-distribution (namely, nuScenes) and out-of-distribution (namely, SemanticKITTI) settings. For representations distilled from CLIP in \cref{tab:clip_baseline}, we observe that the limited U of representations distilled through similarity losses, as opposed to the source, leads to comparatively weaker performance on in-distribution tasks with few-shot learning when compared with contrastive loss. For instance, ST-SLidR representations enhance performance on the nuScenes dataset by +1.5\% compared to $\text{Sim}_{spl}$. Furthermore, by applying relational constraints and without the need for negative samples, we can bridge the U gap. This results in $\text{Rel}_{spl}$ outperforming $\text{Sim}_{spl}$ by +1.59\% and +1.1\% on nuScenes and SemKITTI, respectively. When examining 3D representations distilled from DINOv2 in~\cref{tab:dino_baseline}, we note a source U of 2.8. Here, the similarity loss closely aligns with the source's U more than the contrastive loss. The contrastive loss (PPKT and SLidR) shows significantly poorer performance compared to the similarity loss. Additionally, 3D representations distilled by ST-SLidR attain a U of 3.589 and a T of 0.4326, aligning more closely with the source U and T of 2.8 and 0.51 than those distilled by the SLidR loss. Lastly, we find that relational constraints narrow the G more effectively than similarity losses, thereby enhancing the few-shot performance by +1.05\% and +1.19\% on pixel and superpixel-based losses, respectively, on the nuScenes dataset compared to similarity losses.\\
\begin{figure}[t]
\centering 
\begin{minipage}{.48\textwidth}
\centering
\captionof{table}{Zero-shot segmentation performance of relational loss compared to state-of-the-art methods.}
\label{tab:zero_shot_performance}
\scalebox{0.75}{
\begin{tabular}{ccccc}
\toprule
\multirow{2}{*}{Method} & \multirow{2}{*}{Publication} & \multirow{2}{*}{\begin{tabular}[c]{@{}c@{}}Class Names \\ Required\end{tabular}} & \multirow{2}{*}{\begin{tabular}[c]{@{}c@{}}Uses \\ SAM\end{tabular}} & \multirow{2}{*}{\begin{tabular}[c]{@{}c@{}}3D \\ mIoU\end{tabular}} \\
                        &                              &                                                                                  &                                                                            &                          \\ \hline
MaskCLIP~\cite{zhou2022extractMaskCLIP}                & ECCV 2022                     & \ding{55}                                                                                & \ding{55}                                                                          & 12.80                    \\
OpenScene~\cite{peng2023openscene}               & CVPR 2023                     & \ding{55}                                                                              & \ding{55}                                                                          & 14.60                    \\
CLIP2Scene~\cite{chen2023clip2scene}              & CVPR 2023                     & \ding{51}                                                                               & \ding{55}                                                                          & 20.80                    \\
\text{Rel}$_{pl}$ (ours)          & -                            & \ding{55}                                                                                & \ding{55}                                                                          & 23.53                    \\ \hline
TLF~\cite{chen2023towards}                 & NeurIPS 2023                  & \ding{51}                                                                               & \ding{51}                                                                         & 26.80                    \\
\text{Rel}$_{spl}$ (ours)          & -                            & \ding{55}                                                                                & \ding{51}                                                                         & 26.03    
\\
\bottomrule
\end{tabular}}
\captionof{table}{Effect of relational loss components. Distilling from CLIP image encoder 
 with source U of 1.54 and T of 0.73.}
\label{tab:relation_loss_ablation}
\scalebox{0.72}{
\begin{tabular}{lcccccc}
\toprule
\multicolumn{1}{c}{\multirow{3}{*}{Loss}} & \multicolumn{1}{c}{\multirow{3}{*}{Uniformity}} & \multicolumn{1}{c}{\multirow{3}{*}{Tolerance}} & \multicolumn{1}{c}{\multirow{3}{*}{\begin{tabular}[c]{@{}c@{}}Modality \\ Gap\end{tabular}}} & \multicolumn{2}{c}{nuScenes}                                                                                                                                                                                               & \multicolumn{1}{c}{KITTI}                                                             \\ \cline{5-7} 
\multicolumn{1}{c}{}                      & \multicolumn{1}{c}{}                            & \multicolumn{1}{c}{}                           & \multicolumn{1}{c}{}                                                                         & \multicolumn{1}{c}{\multirow{2}{*}{ZS}} & \multicolumn{1}{c}{\multirow{2}{*}{\begin{tabular}[c]{@{}c@{}}Ft \\ 1\%\end{tabular}}} & \multicolumn{1}{c}{\multirow{2}{*}{\begin{tabular}[c]{@{}c@{}}Ft\\ 1\%\end{tabular}}} \\
\multicolumn{1}{c}{}                      & \multicolumn{1}{c}{}                            & \multicolumn{1}{c}{}                           & \multicolumn{1}{c}{}                                                                         & \multicolumn{1}{c}{}                    & \multicolumn{1}{c}{}                                                                   & \multicolumn{1}{c}{}                                                                  \\ \hline
$\text{Sim}_{pl}$                                   & 0.5519                                          & 0.9477                                         & 0.00045                                                                                      & 20.84                                   & 44.39                                                                                  & 45.59                                                                                 \\
+cross                                    & 0.6426                                          & 0.9386                                         & 0.00042                                                                                      & 21.39                                  & \textbf{45.95}                                                                                  & 45.45                                                                                 \\
+intra                                 & 0.9089                                          & 0.9145                                         & \textbf{0.00033}                                                                                      & \textbf{23.53}                                   & 45.67                                                                                  & \textbf{46.06}                                                                                 \\ \hline
$\text{Sim}_{spl}$                                   & 0.5919                                          & 0.9333                                         & 0.00040                                                                                       & 23.93                                   & 45.63                                                                                  & 46.42                                                                                 \\
+cross                                    & 0.7656                                          & 0.9086                                         & 0.00033                                                                                      & 25.05                                   & 46.18                                                                                  & 46.99                                                                                 \\
+intra                                 & 1.0990                                           & 0.8700                                           & \textbf{0.00019}                                                                                      & \textbf{26.03}                                   & \textbf{47.22}                                                                                  & \textbf{47.52}   \\
\bottomrule
\end{tabular}
}
\end{minipage}%
\hfill  
\begin{minipage}{.48\textwidth}
    \centering
    \captionsetup{font=footnotesize}
    \captionof{table}{Improvement of finetuned models with respect to majority versus minority classes. Similar to ST-SLidR~\cite{stcl},  We group classes based on whether
their superpixels occupy more than 5\% of the superpixels in nuScenes training set.}
    \label{tab: minority_majority}
    \setlength{\tabcolsep}{0.5em}\scalebox{0.9}{
    \begin{tabular}{cccc}
\toprule
\multirow{2}{*}{\begin{tabular}[c]{@{}c@{}}2D \\ Encoder\end{tabular}} & \multicolumn{1}{c}{\multirow{2}{*}{Loss}} & \multicolumn{1}{c}{\multirow{2}{*}{\begin{tabular}[c]{@{}c@{}}Majority\\  (mIoU)\end{tabular}}} & \multicolumn{1}{c}{\multirow{2}{*}{\begin{tabular}[c]{@{}c@{}}Minority \\ (mIoU)\end{tabular}}} \\
                                                                       & \multicolumn{1}{c}{}                      & \multicolumn{1}{c}{}                                                                            & \multicolumn{1}{c}{}                                                                            \\ \hline
\multicolumn{1}{c|}{\multirow{6}{*}{CLIP}}                             & $\text{Sim}_{pl}$                                    & 69.16                                                                                           & 33.35                                                                                           \\
\multicolumn{1}{c|}{}                                                  & $\text{Rel}_{pl}$                                    & \textbf{69.59}                                                                                           & \textbf{34.80}                                                                                           \\
\multicolumn{1}{c|}{}                                                  & {\fontfamily{ppl}\selectfont  \textit{Gain}}                                      & {\fontfamily{ppl}\selectfont \textit{+0.43}}                                                                                            & {\fontfamily{ppl}\selectfont \textit{+1.45}}                                                                                            \\ \cline{2-4} 
\multicolumn{1}{c|}{}                                                  & $\text{Sim}_{spl}$                                    & 67.58                                                                                           & 36.06                                                                                           \\
\multicolumn{1}{c|}{}                                                  & $\text{Rel}_{spl}$                                    & \textbf{67.75}                                                                                           & \textbf{38.77}                                                                                           \\
\multicolumn{1}{c|}{}                                                  & {\fontfamily{ppl}\selectfont  \textit{Gain}}                                      & {\fontfamily{ppl}\selectfont \textit{+0.17}}                                                                                            & {\fontfamily{ppl}\selectfont \textit{+2.71}}                                                                                            \\ \hline
\multicolumn{1}{c|}{\multirow{6}{*}{DINOv2}}                           & $\text{Sim}_{pl}$                                    & 70.89                                                                                           & 34.24                                                                                           \\
\multicolumn{1}{c|}{}                                                  & $\text{Rel}_{pl}$                                   & \textbf{71.22}                                                                                           & \textbf{35.85}                                                                                           \\
\multicolumn{1}{c|}{}                                                  & {\fontfamily{ppl}\selectfont \textit{Gain}}                                      & {\fontfamily{ppl}\selectfont \textit{+0.33}}                                                                                            & {\fontfamily{ppl}\selectfont \textit{+1.61}}                                                                                            \\ \cline{2-4} 
\multicolumn{1}{c|}{}                                                  & $\text{Sim}_{spl}$                                    & 68.15                                                                                           & 38.53                                                                                           \\
\multicolumn{1}{c|}{}                                                  & $\text{Rel}_{spl}$                                    & \textbf{68.80}                                                                                           & \textbf{40.20}                                                                                           \\
\multicolumn{1}{c|}{}                                                  & {\fontfamily{ppl}\selectfont \textit{Gain}}                                      & {\fontfamily{ppl}\selectfont\textit{+0.65}}                                                                                            & {\fontfamily{ppl}\selectfont \textit{+1.67}}      \\
\bottomrule
\end{tabular}}
\end{minipage}
\end{figure}
\noindent\textbf{CLIP Backbones} \cref{tab:clip_arch} depicts the performance of contrastive, similarity, and relational losses when distilling from different CLIP backbones. Relational losses achieve the lowest modality gap, surpassing both contrastive and similarity loss across all 2D encoders. In zero-shot tasks, relational losses either match or exceed the performance of similarity losses, whereas contrastive distillation consistently underperforms. The performance of contrastive distillation on few-shot tasks proves unpredictable. For instance, distilling from ViT-B32 using similarity loss $\text{Sim}_spl$ surpasses distilling using contrastive loss SLidR in mIOU by +1.45\% and +1.87\% on nuScenes and SemanticKITTI, respectively. Conversely, distilling from ViT-L14, SLidR achieves competitive performance, outperforming similarity loss $\text{Sim}_{spl}$ by +1.58\% and +1.57\% on the same datasets. Notably, relational losses bridge the gap in few-shot learning while maintaining robust zero-shot task performance. This indicates that relational losses utilize inherent 2D representation relationships, avoiding ungrounded negative samples common in contrastive loss, which facilitates learning representations more aligned with the source's U, thereby enhancing few-shot task performance. \\
\noindent\textbf{Class Imbalance} We investigate the performance of finetuned models on semantic segmentation using 3D representations distilled using similarity and relational losses. Similar to ST-SLidR~\cite{stcl}, we distinguish between majority and minority classes based on the percentage of superpixels they occupy in the nuScenes dataset. The 11 classes occupying less than 5\% of the superpixels are considered to be in the minority set, while the remaining classes are in the majority set. In~\cref{tab: minority_majority}, compared to similarity losses, relational losses learn representations that significantly improve performance on minority classes for pixel and superpixel-based losses without degrading performance on majority classes. 

\noindent\textbf{Relational Loss Ablation} We investigate the contribution of cross-modal and intra-modal constraints for pixel-based and superpixel-based losses. In \cref{tab:relation_loss_ablation}, we observe that both constraints lead to learning 3D representations that are closer to the source U and T, and thus result in a lower G compared to similarity losses. Moreover, both constraints lead to improved performance on zero-shot and few-shot segmentation tasks. Interestingly, the superpixel-based relational loss with both constraints (last row) results in the smallest gap in U and T compared to other losses, leading to the best performance on semantic segmentation.

\section{Conclusion}
\label{sec:conclusion}
In this work, we study the impact of the state-of-the-art 2D-to-3D distillation frameworks when applied to AD datasets on the structure of the learned 3D representation. We reveal a significant structural gap between the 2D and the 3D representations and show that this gap is negatively correlated with the utility of the learned 3D representations for solving 3D zero-shot and few-shot segmentation tasks. Our proposed relational loss bridges this structural gap, resulting in well-aligned 3D representations that outperform representations learned via contrastive loss on zero-shot segmentation tasks. In addition, compared to the similarity loss, our relational loss results in 3D representations that consistently improve in-distribution and out-of-distribution few-shot segmentation tasks.

%
%
\bibliographystyle{splncs04}
\bibliography{egbib}

\newpage
\appendix
\section{Toy Example: Learning on a Unit Sphere}
\label{app:toy_example}
\subsection{Setup}
\begin{figure}
    \centering
    \includegraphics[width=\textwidth]{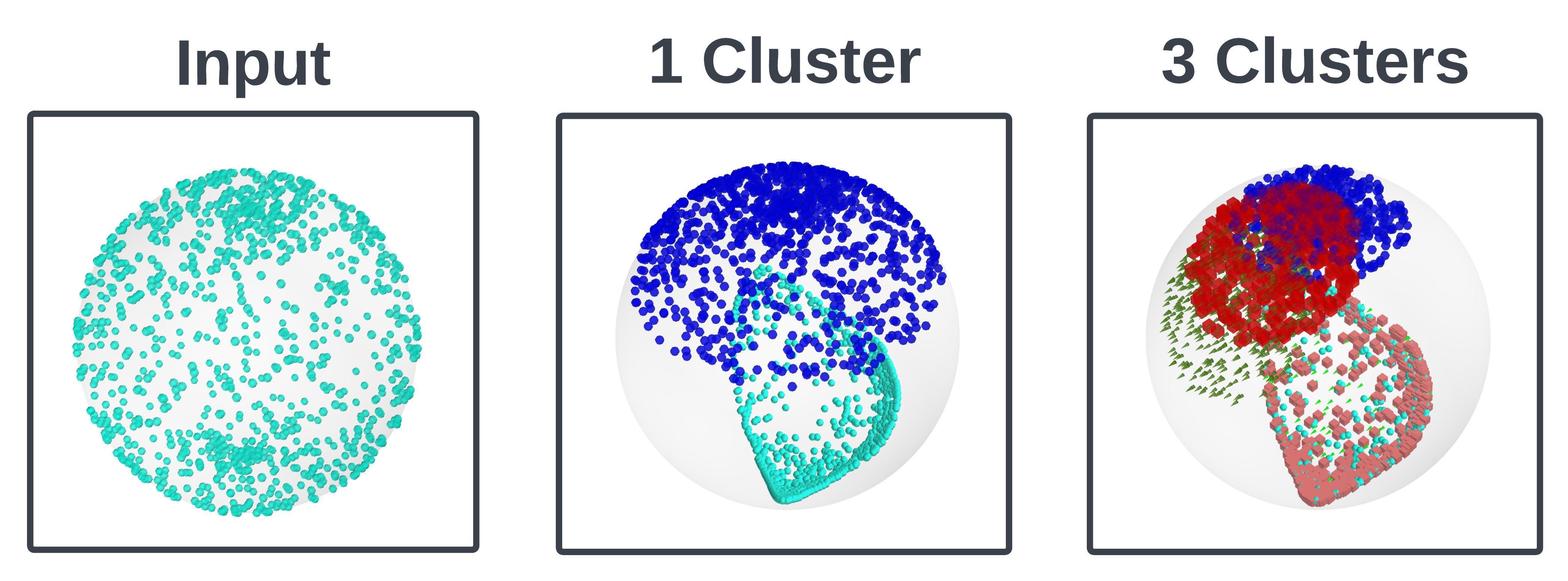}
    \caption{\textbf{Left:} Input points to the MLP model in both the $3$ cluster and $1$ cluster setups. \textbf{Middle:} The 1 cluster setup of the toy example, with the output of the randomly initialized MLP in cyan and the target cluster in blue. \textbf{Right:} The 3 cluster setup, with the output of the randomly initialized MLP in cyan, light green, and pink and the target cluster in blue, green, and red.}
    \label{fig:toy example setup.}
\end{figure}

In this section, we provide additional details about the toy example in~\cref{fig:toy_example}. We start with $1000$ or $1500$ (depending on the setting below) uniformly distributed points over a 3D unit sphere, representing point features before the distillation phase. This set of uniform input points can be visualized as the leftmost plot in~\cref{fig:toy example setup.}. Using a 2-layer MLP, we learn to align each input point with a point from the source 3D space by directly predicting the output points' normalized $x,y,z$ location. The MLP is a simple sequential model with the following layers; 3x512 Linear layer, ReLU, 512x3 Linear layer. 
\begin{figure}
    \centering
    \includegraphics[width=\textwidth]{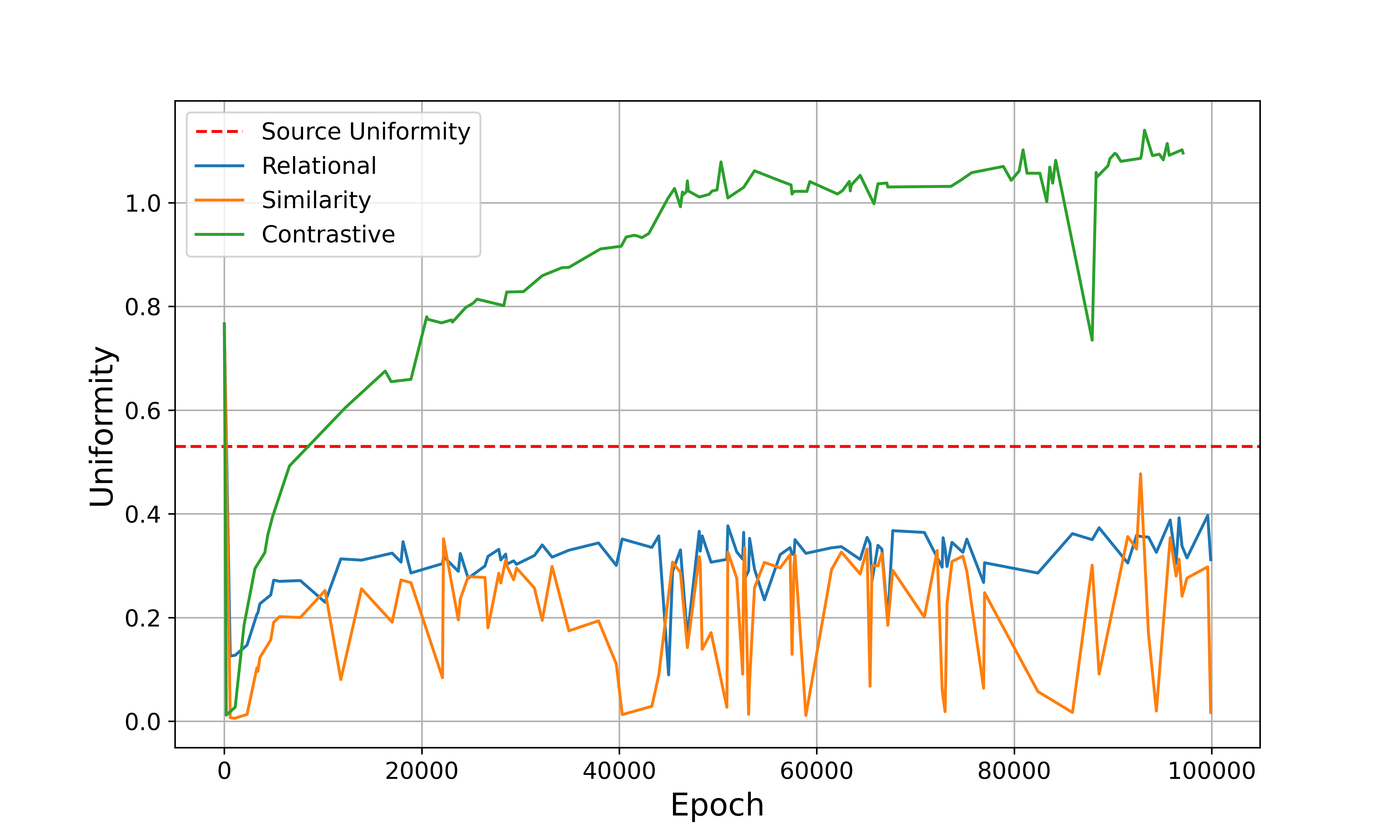}
    \caption{The uniformity values achieved by each loss in comparison to the uniformity of source 3D space as training progresses on the 3-cluster setup.}
    \label{fig:training_uniformity}
\end{figure}

\begin{figure}
    \centering
    \includegraphics[width=\textwidth]{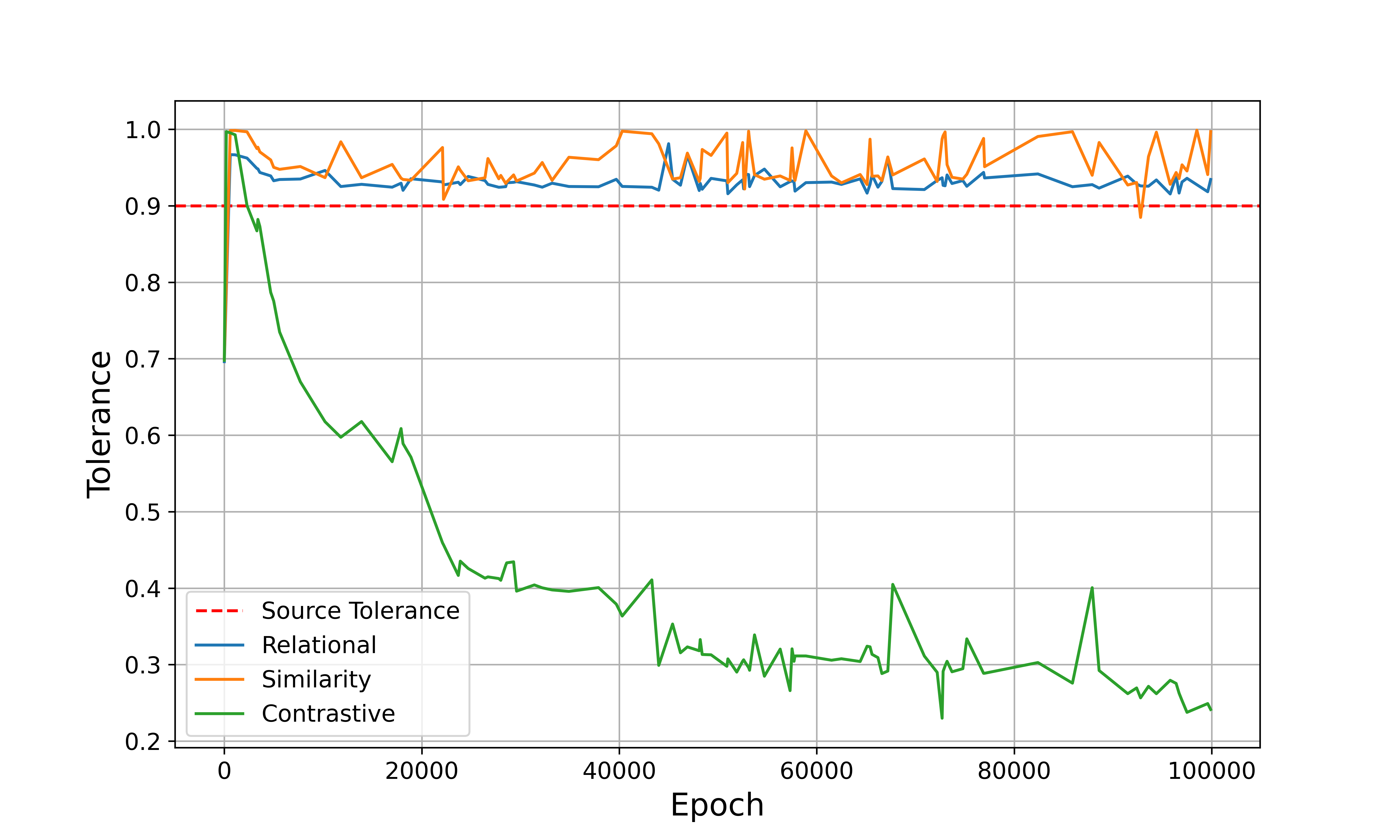}
    \caption{The tolerance values achieved by each loss in comparison to the tolerance of source 3D space as training progresses on the 3-cluster setup.}
    \label{fig:training_tolerance}
\end{figure}
We try this toy example with 2 source 3D spaces, a single cluster of $1000$ points shown in the middle plot of ~\cref{fig:toy example setup.}, and a harder setting of three clusters of $500$ points, randomly sampled similarly, but with a shift in angle across the surface of the hypersphere (~\cref{fig:toy example setup.}, right). Both plots also visualize the MLP projection of the input 3D space at the beginning of training (step 0). The single-cluster setup and the three-cluster setup are trained for 50000 and 100000 epochs respectively, using ADAM and a learning rate of 0.0001. We train with a batch size of $1000$ and $1500$ respectively, meaning that all the points are considered at every training step (epoch and step are equal). For the contrastive loss, we tried 3 temperatures (1.0, 0.1, 0.01) and chose the best results, which consistently were at 0.1 temperature. 

\subsection{Results}
We show the results of learning on this toy example with the Contrastive, Similarity, and Relational losses on the two settings in Table~\ref{tab:toy_example_results}. We notice that the Relational loss produces a predicted 3D space that is closest in Uniformity and Tolerance to the source 3D space in both settings. The Relational loss also produces the lowest modality gap between the two when compared to the Similarity and Contrastive losses.
\begin{table}[h]
    \centering
    \begin{tabular}{c c c c c}
         \textbf{Experiment} & \textbf{Loss} & $\Delta$ \textbf{U} & $\Delta$ \textbf{T} & \textbf{G}  \\ \midrule
         \multirow{3}{*}{1 Cluster}  & Contrastive & 0.26  & 0.48 & 1.28\\
           & Similarity & 0.41  &  0.17& 0.11 \\
             & Relational (Ours) &  \textbf{0.21} & \textbf{0.11 }& \textbf{0.08} \\ \midrule
           \multirow{3}{*}{3 Clusters}  & Contrastive &  0.49 & 0.64 &  1.38\\
           & Similarity &  0.32 & 0.05 & 0.08 \\
             & Relational (Ours)  & \textbf{0.22}   & \textbf{0.01} &  \textbf{0.05}\\ \bottomrule
    \end{tabular}
    \caption{The results of the structural difference between the source 3D space and the predicted 3D space from the Contrastive, Similarity, and Relational losses on both the 1 and 3 cluster settings.}
    \label{tab:toy_example_results}
\end{table}

We also present one additional set of interesting results that highlight the effectiveness of our proposed Relational loss. Figures~\ref{fig:training_uniformity} and~\ref{fig:training_tolerance} show the evolution of the uniformity and tolerance of the output 3D space \footnote{3-cluster setup, same observations were made in the 1-cluster setup as well.} in comparison to the source 3D space, as training progresses from 0 epochs to 100,000 epochs. The results for the Contrastive loss are clear, it substantially increases uniformity to a very high value at the expense of tolerance. However, what is interesting is that the Relational loss consistently produces uniformity and tolerance values for the 3D output space that is closer to the source 3D space than those of the Similarity loss. This means that in the case of premature/overdue termination or other inefficiencies in training, the Relational loss has a much higher probability of outputting a 3D space closer to the source 3D space when compared to the similarity loss.  
\section{Training and Inference Details}
\label{app:training}
 For point cloud data augmentation, we apply linear transformations to the point cloud which include random rotations around $z$-axis and flipping around $x$-axis and $y$-axis. In addition, we also randomly select a cube and drop all points within the cube~\cite{Zhang_2021_depthcontrast}. For 2D augmentations, we apply random crop-resize while ensuring a minimum number of 3D points exist in the cropped scene~\cite{sautier2022slidr}. We distill using pixel-based and superpixel-based losses. For superpixel-based losses, we prompt SAM~\cite{Kirillov_2023_ICCV_sam} to generate superpixel masks for each image~\cite{liu2023seal}. Augmentations applied to images are also applied to superpixel masks. We pre-train the point cloud encoder and the projection layer, for 20 epochs on 2 A100 GPUs with a batch size of 8. Similar to previous works~\cite{sautier2022slidr, stcl}, we employ an SGD optimizer with a 0.9 momentum, a cosine annealing learning rate scheduler and an initial learning rate of 0.5. Lastly, for regularization purposes, we implement a weight decay of 0.0001 and a dampening factor of 0.1. During the pre-training phase, the vision encoders are frozen and gradients only propagate through the 3D point encoder. \\
 For zero-shot 3D semantic segmentation on the nuScenes dataset, we utilize the prompt template proposed in~\cite{puy2023revisit} to compute the CLIP text embeddings for nuScenes classes.
 \section{Projection Layers}
\label{app:proj_layers}
Since we are interested in fine-grained semantic understanding of the scene, we preserve dense visual features by removing the last attention pooling layer of CLIP models and applying a projection layer to input tokens similar to MaskCLIP~\cite{zhou2022extractMaskCLIP}. Unlike SLidR~\cite{sautier2022slidr}, we are not only interested in few-shot segmentation but also cross-modal retrieval using language prompts during inference (i.e., zero-shot segmentation). Therefore, similar to~\cite{peng2023openscene, puy2023revisit}, we do not apply any projection layers to the output of the vision encoders. For CLIP and DINOv2, we interpolate positional embeddings to allow the distillation of crops of images with different resolutions. For the 3D backbone, to enable distilling from vision encoders with different output dimensions, we learn a projection layer that maps the output of the point encoder to the output of the vision encoder.
\section{Effect of $\tau$ on Contrastive Losses}
\label{app:tao_effect_on_contrastive}
Contrastive losses (CL) learn by pulling positive samples while pushing against all negative samples in the current batch. As demonstrated in~\cite{wang2021understanding}, the behaviour of the CL changes based on the temperature scaling parameter denoted by $\tau$. On one hand, a low temperature leads to amplifying the relative scale of the gradients of the closest negative samples to the positive samples. This results in a highly uniform representation at the expense of learning semantically coherent clusters. On the other hand, a high temperature leads to a more uniform magnitude of the gradient across all negative samples, which results in semantically coherent clusters at the expense of learning a representation with low uniformity. Due to the abundance of self-similarity in autonomous driving datasets, many of the negative samples in the current batch are false negatives~\cite{stcl} as they belong to the same semantic class as the positive sample. Here, we would like to study the effect of self-similarity on the behaviour of the contrastive loss by varying $\tau$ in the range from 0.07 to 1.0. Looking at~\cref{tab:tao_effect_on_contrastive}, we observe that increasing $\tau$ reduces the uniformity, and increases the tolerance of the distilled 3D representation. More importantly, increasing $\tau$, minimizes the scale of the gradients from the negative samples that are close to the positive sample. In self-similar environments, these negative samples are probably from the same semantic class as the positive sample, thus avoiding pushing against these samples leads to a significant improvement in zero-shot mIoU from 14.53\% to 18.26\% as we increase $\tau$ from 0.07 to 1.0. However, increasing $\tau$ comes at the expense of increasing the modality gap. Increasing $\tau$ from 0.07 to 1.0, results in a significant drop in mIoU in the few-shot segmentation setting; 45.31$\%$ to 41.08$\%$ and 45.77$\%$ to 44.18$\%$ on nuScenes and semantic KITTI respectively. We reason that contrary to the effectiveness of CL in general cross-modal distillation settings~\cite{girdhar2023imagebind}, CL learns sub-optimal representations in environments with abundance of self-similarity like autonomous driving scenes. 
\section{Zero-shot Visualizations}
\label{app:zero_shot_vis}
\begin{figure}
    \centering
    \includegraphics[width=\textwidth]{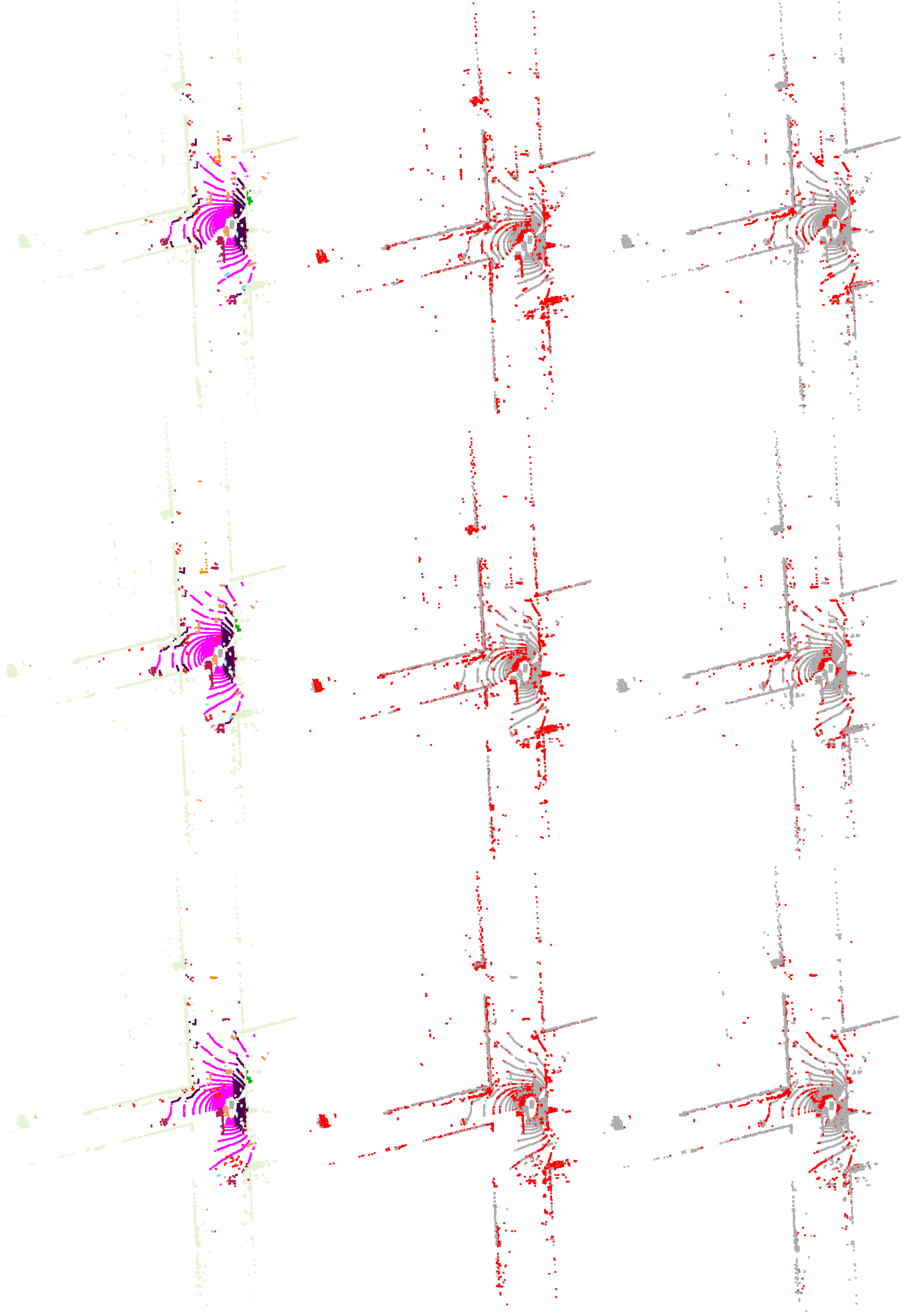}
    \caption{Visualization of the error maps of zero-shot predictions using 3D models pre-trained using contrastive and relational losses on nuscenes dataset. Here, gray and red points indicate correct and wrong predictions respectively. \textbf{Left:} Ground-truth Labels \textbf{Middle:} Contrastive Loss Error Map \textbf{Right:} Relational Loss Error Map}
    \label{fig:train_error_map_scene0239}
\end{figure}

\begin{figure}
    \centering
    \includegraphics[width=0.75\textwidth]{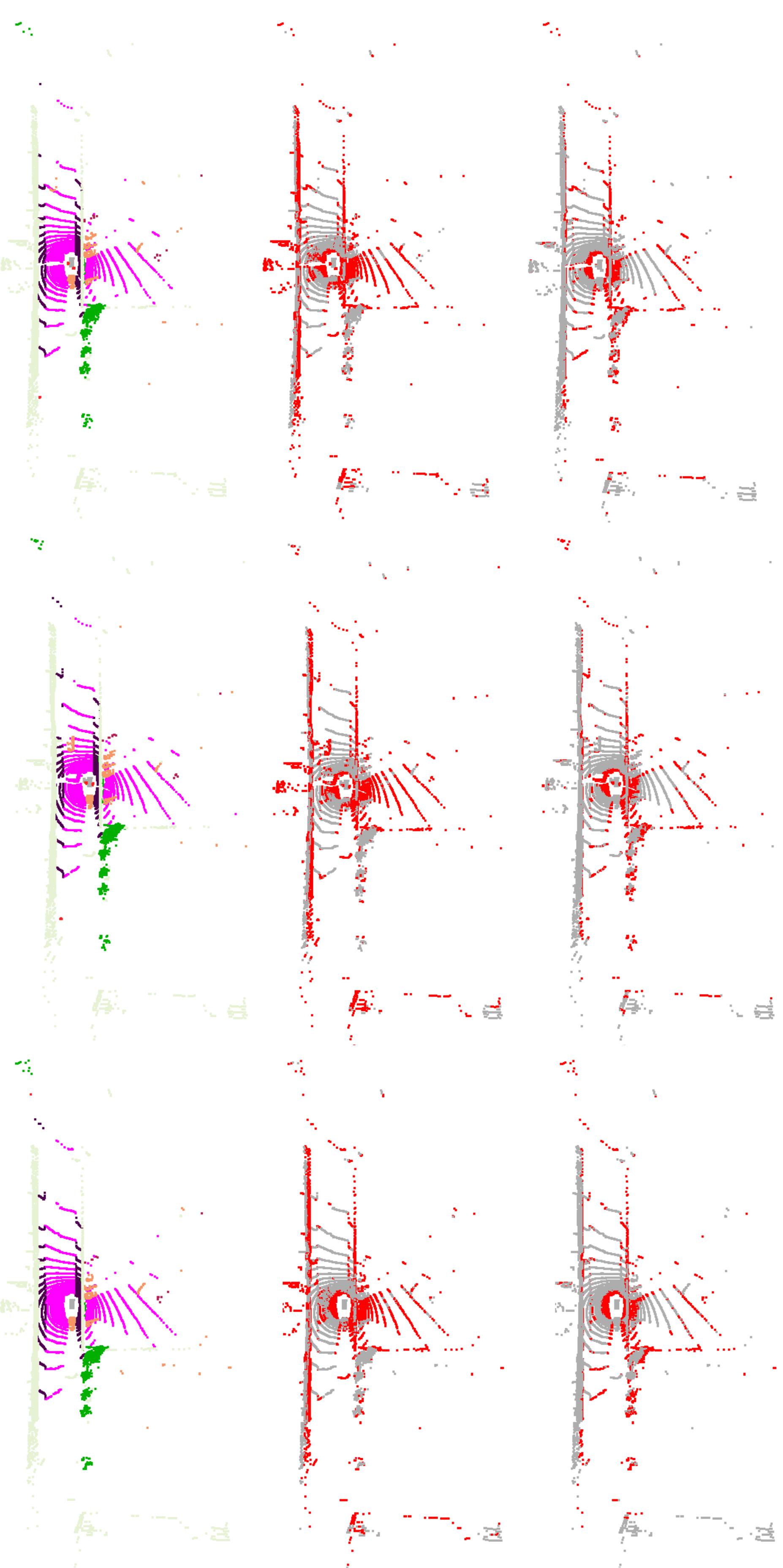}
    \caption{Visualization of the error maps of zero-shot predictions using 3D models pre-trained using contrastive and relational losses on nuscenes dataset. Here, gray and red points indicate correct and wrong predictions respectively. \textbf{Left:} Ground-truth Labels \textbf{Middle:} Contrastive Loss Error Map \textbf{Right:} Relational Loss Error Map}
    \label{fig:val_error_map_scene0777}
\end{figure}

In~\cref{fig:train_error_map_scene0239} and~\cref{fig:val_error_map_scene0777}, we visualize zero-shot predictions on nuScenes dataset by distilling CLIP 2D representations using models pre-trained using superpixel-based contrastive and relational losses. We observe that contrastive losses show significantly more errors that relational losses as depicted by the points labelled in red. Due to the abundance of self-similarity in autonomous driving data, coupled with the hardness-aware property of contrastive losses, the learned point features are not aligned with CLIP text features. Our observations do not suggest that contrastive loss is unsuitable for distillation tasks. Rather, in contexts where self-similarity is not a concern, contrastive loss has been demonstrated to perform effectively in distillation tasks, as evidenced by the findings in ~\cite{girdhar2023imagebind}.
\section{Finetuning Per-class Performance}
\label{app:per_task_performance}
We present the per-class intersection-over-union for the task of semantic segmentation averaged over 3 runs. The point encoders are pre-trained using superpixel-based relational losses, and then fine-tuned on 1\% of nuScenes dataset. We depict the results of distilling from CLIP and DINOv2 2D encoders in~\cref{tab:iou_per_class_superpixeldriven_clip} and~\cref{tab:iou_per_class_superpixeldriven_dinov2} respectively. Classes belonging to minority set (i.e., represent less than 5\% of the superpixels across the images of nuScenes training set) are bolded. 
\section{Limitations}
We have demonstrated that the relational constraints can bridge the structural gap between 2D and the distilled 3D representations leading to more generalizable features for downstream tasks. These constraints are supervised by  computing similarities between image features in a batch. We hypothesize that not all relations are equally useful and not all relations are accurate. This is especially true for CLIP vision encoders which have poor localization information at the pixel-level~\cite{zhou2022extractMaskCLIP}. Therefore, it would be useful to investigate ways to detect outlier image features which would lead to more accurate relational constraints.
In addition, to extract point to pixel correspondences (i.e., positive pairs), image-to-2D distillation frameworks assume accurate camera-LiDAR calibration information. While superpixel-based losses are more robust to calibration errors compared to pixel-based losses, large errors in calibration can limit the utility of 2D-to-3D distillation frameworks. 


\begin{table}[]
\centering
\begin{tabular}{crrrccc}
\toprule
\multicolumn{1}{c}{\multirow{3}{*}{\begin{tabular}[c]{@{}c@{}}$\tau$\end{tabular}}} & \multicolumn{1}{c}{\multirow{3}{*}{Uniformity}} & \multicolumn{1}{c}{\multirow{3}{*}{Tolerance}} & \multicolumn{1}{c}{\multirow{3}{*}{\begin{tabular}[c]{@{}c@{}}Modality \\ Gap\end{tabular}}} & \multicolumn{2}{c}{nuScenes}                                                                           & KITTI                                                                     \\ \cline{5-7} 
\multicolumn{1}{c}{}                                                                             & \multicolumn{1}{c}{}                            & \multicolumn{1}{c}{}                           & \multicolumn{1}{c}{}                                                                         & \multirow{2}{*}{Zero-shot} & \multirow{2}{*}{\begin{tabular}[c]{@{}c@{}}Finetuning\\ 1\%\end{tabular}} & \multirow{2}{*}{\begin{tabular}[c]{@{}c@{}}Finetuning\\ 1\%\end{tabular}} \\
\multicolumn{1}{c}{}                                                                             & \multicolumn{1}{c}{}                            & \multicolumn{1}{c}{}                           & \multicolumn{1}{c}{}                                                                         &                            &                                                                           &                                                                           \\ \hline
0.07                                                                                             & 3.52                                            & 0.5217                                         & 0.00158                                                                                      & 14.53                      & 45.31                                                                     & 45.77                                                                     \\
0.1                                                                                              & 3.47                                            & 0.5514                                         & 0.00158                                                                                     & 15.74                      & 45.42                                                                     & 45.62                                                                     \\
0.2                                                                                              & 3.33                                            & 0.6031                                         & 0.00178                                                                                      & 17.25                      & $\mathbf{45.42}$                                                                     & $\mathbf{46.05}$                                                                     \\
0.5                                                                                              & 3.17                                            & 0.6202                                         & 0.00184                                                                                      & $\mathbf{18.52}$                      & 42.96                                                                     & 44.18                                                                     \\
1.0                                                                                              & 3.13                                            & 0.6358                                         & 0.00178                                                                                      & 18.26                      & 41.08                                                                     & 44.18                                                                    \\
\bottomrule
\end{tabular}
\caption{ The effect of the temperature scaling parameter denoted by $\tau$ in~\cref{eq:loss_contrastive} on the structure of the 3D distilled representations from CLIP and its effect on zero-shot and few-shot semantic segmentation.} 
\label{tab:tao_effect_on_contrastive}
\vspace{-50pt}
\end{table}
\begin{table}[h]
\centering
\resizebox{\textwidth}{!}{
\begin{tabular}{lcccccccccccccccc}
\toprule
Method      & mIoU & \rotatebox{90}{\textbf{bicycle}} & \rotatebox{90}{\textbf{bus}}  & \rotatebox{90}{\textbf{car}}  & \rotatebox{90}{\textbf{const. veh.}} & \rotatebox{90}{\textbf{motorcycle}} & \rotatebox{90}{\textbf{pedestrian}} & \rotatebox{90}{\textbf{traffic cone}} & \rotatebox{90}{\textbf{trailer}} & \rotatebox{90}{\textbf{truck}} & \rotatebox{90}{driv. surf.} & \rotatebox{90}{other flat} & \rotatebox{90}{sidewalk} & \rotatebox{90}{terrain} & \rotatebox{90}{manmade} & \rotatebox{90}{vegetation} \\ \hline
Random      & 30.3     & 0.0     & 8.1  & 65.0 & 0.1         & 6.6        & 21.0       & 9.0          & 9.3     & 25.8  & 89.5        & 14.8       & 41.7     & 48.7    & 72.4    & 73.3       \\
$\text{Sim}_{spl}$       & 45.6     & 6.0     & 38.3 & 75.0 & 11.1         & 25.3       & 55.4       & 41.8         & 25.1    & 46.6  & 91.6        & 37.1       & 52.8     & 60.2    & 81.4    & 82.4      \\
$\text{Rel}_{spl}$    & \textbf{47.2}     & 6.1     & 50.4 & 74.3 & 16.3         & 32.8       & 56.5      & 43.9         & 24.1   & 44.5  & 92.0        & 37.4  & 53.0    & 59.8    & 81.7   & 82.6    \\ \hline
{\fontfamily{ppl}\selectfont\textit{Gain}} & {\fontfamily{ppl}\selectfont\textit{+1.6}}      & {\fontfamily{ppl}\selectfont\textit{+0.1}}     & {\fontfamily{ppl}\selectfont\textit{+12.0}}  & {\fontfamily{ppl}\selectfont\textit{-0.7}}  & {\fontfamily{ppl}\selectfont\textit{+5.2}}         & {\fontfamily{ppl}\selectfont\textit{+7.5}}        & {\fontfamily{ppl}\selectfont\textit{+1.1}}        & {\fontfamily{ppl}\selectfont\textit{+2.1}}          & {\fontfamily{ppl}\selectfont\textit{-1.0}}     & {\fontfamily{ppl}\selectfont\textit{-2.1}}   & {\fontfamily{ppl}\selectfont\textit{+0.4}}         & {\fontfamily{ppl}\selectfont\textit{+0.3}}        & {\fontfamily{ppl}\selectfont\textit{+0.2}}      & {\fontfamily{ppl}\selectfont\textit{-0.4}}     & {\fontfamily{ppl}\selectfont\textit{+0.3}}     & {\fontfamily{ppl}\selectfont\textit{+0.2}}        \\
\bottomrule
\end{tabular}}
\caption{3D semantic segmentation using 1\% of labelled data for fine-tuning on nuscenes dataset on official validation set. We report the mean performance of 3 pretrained models distilled from CLIP ViT-B16 using superpixel-driven losses; $\text{Sim}_{spl}$ and $\text{Rel}_{spl}$. Minority classes are bolded and gain is reported relative to $\text{Sim}_{spl}$} \label{tab:iou_per_class_superpixeldriven_clip}
\vspace{-40pt}
\end{table}
\begin{table}[h]
\centering
\resizebox{\textwidth}{!}{
\begin{tabular}{lccccccccccccccccc}
\toprule
Method      & mIoU & \rotatebox{90}{\textbf{bicycle}} & \rotatebox{90}{\textbf{bus}}  & \rotatebox{90}{\textbf{car}}  & \rotatebox{90}{\textbf{const. veh.}} & \rotatebox{90}{\textbf{motorcycle}} & \rotatebox{90}{\textbf{pedestrian}} & \rotatebox{90}{\textbf{traffic cone}} & \rotatebox{90}{\textbf{trailer}} & \rotatebox{90}{truck} & \rotatebox{90}{driv. surf.} & \rotatebox{90}{other flat} & \rotatebox{90}{sidewalk} & \rotatebox{90}{terrain} & \rotatebox{90}{manmade} & \rotatebox{90}{vegetation} \\ \hline
Random      & 30.3     & 0.0     & 8.1  & 65.0 & 0.1      & 6.6        & 21.0       & 9.0          & 9.3     & 25.8  & 89.5        & 14.8       & 41.7     & 48.7    & 72.4    & 73.3       \\
$\text{Sim}_{spl}$       &47.2          & 6.5     & 48.3  & 74.4  & 24.1    & 35.3       & 61.9      &33.7      &21.1   &41.6   &92.6      &35.0   &55.2    &60.6    &82.1   &83.6      \\
$\text{Rel}_{spl}$    &\textbf{48.4}      & 7.8     & 55.3  & 76.0  & 24.8    & 32.6       & 61.2      &40.7      &21.9   & 41.5  &92.8      &36.4   &55.9    &61.0    &82.7   &83.9    \\ \hline
{\fontfamily{ppl}\selectfont\textit{Gain}} & {\fontfamily{ppl}\selectfont\textit{+1.2}}    & {\fontfamily{ppl}\selectfont\textit{+1.3}}     & {\fontfamily{ppl}\selectfont\textit{+7.0}}  & {\fontfamily{ppl}\selectfont\textit{+1.6}}  & {\fontfamily{ppl}\selectfont\textit{+0.7}}         & {\fontfamily{ppl}\selectfont\textit{-2.7}}        & {\fontfamily{ppl}\selectfont\textit{-0.7}}        & {\fontfamily{ppl}\selectfont\textit{+7.0}}          & {\fontfamily{ppl}\selectfont\textit{+0.8}}     & {\fontfamily{ppl}\selectfont\textit{-0.1}}   & {\fontfamily{ppl}\selectfont\textit{+0.2}}         & {\fontfamily{ppl}\selectfont\textit{+1.4}}        & {\fontfamily{ppl}\selectfont\textit{+0.7}}      & {\fontfamily{ppl}\selectfont\textit{+0.4}}     & {\fontfamily{ppl}\selectfont\textit{+0.6}}     & {\fontfamily{ppl}\selectfont\textit{+0.3}}        \\
\bottomrule
\end{tabular}}
\caption{3D semantic segmentation using 1\% of labelled data for fine-tuning on nuscenes dataset on official validation set. We report the mean performance of 3 pretrained models distilled from DINOv2 using superpixel-driven losses; $\text{Sim}_{spl}$ and $\text{Rel}_{spl}$.} \label{tab:iou_per_class_superpixeldriven_dinov2}
\vspace{-30pt}
\end{table}

\end{document}